\documentclass[10pt,a4paper]{article}
\usepackage[margin=22mm]{geometry}
\usepackage{lmodern,microtype}
\usepackage[numbers,sort&compress]{natbib}
\usepackage[T1]{fontenc}
\usepackage{amsmath,amssymb,booktabs,graphicx,tabularx,array,needspace}
\usepackage{pgfplots,xurl}
\usepackage[hidelinks]{hyperref}
\hypersetup{pdftitle={Observability limits of nonrigidity screening in structure from motion},
  pdfauthor={Íñigo Cirauqui-Viloria}}
\usetikzlibrary{arrows.meta,positioning,shapes.geometric}
\pgfplotsset{compat=1.18}
\newcommand{\ResourcePreprintID}{2610.35621v1}

\newcommand{\ScreeningSourceVersion}{0.1.2}
\newcommand{\ScreeningSourceRevision}{3674a4856b9b4896969b22f2deeba1cbacbffa0b}

\newcommand{\ScreeningInputVersion}{0.1.0}

\newcommand{\ScreeningInputReference}{\url{https://doi.org/10.57967/hf/11298}}

\newcommand{\ScreeningMvaVersion}{0.2.0}

\newcommand{\ScreeningMvaRevision}{696a1c17e3841b8069f496950c2fc64f2f19d24c}
\newcommand{\ScreeningMvaReference}{\url{https://doi.org/10.57967/hf/11848}}
\newcommand{\ScreeningMvaRevisionLink}{\href{https://huggingface.co/datasets/icirauqui/sfm-nonrigidity-screening-reproduction/tree/\ScreeningMvaRevision}{\texttt{696a1c1}}}

\ifdefined\pdfgentounicode
  \InputIfFileExists{glyphtounicode.tex}{\pdfgentounicode=1}{}
\fi
\ifdefined\pdftrailerid\pdftrailerid{}\fi
\title{Observability limits of nonrigidity screening in structure from motion}
\author{\'I\~nigo Cirauqui-Viloria\\
\small Universidad de Zaragoza, Zaragoza, Spain\\
\small \href{mailto:icirauqui@gmail.com}{icirauqui@gmail.com}}
\date{}
\begin{document}
\twocolumn[{\begin{@twocolumnfalse}
\maketitle
\begin{abstract}
A static structure-from-motion model can fit every image track even when the scene changes shape. This is possible because motion along a monocular viewing ray leaves the projected point unchanged. A simultaneous second view with sufficient parallax can distinguish the two states. We ask how much evidence of deformation can be extracted from a single static reconstruction and its image tracks. We quantify the image response that remains after removing the effects of small point and camera adjustments and time-varying affine fields, then measure how much of that response aligns with the patterns tested by the detector. For a given motion direction, fixed geometry and known independent Gaussian pixel noise, this analysis gives the displacement amplitude needed for 80\% detection probability at a 5\% false-alarm rate. We simulated a 90-dimensional family of localized motions using eleven measured tube geometries. After nuisance removal, 66 directions remained observable to first order, of which the modal test retained 36. Retaining observable directions in rank-deficient tracks allows two previously rejected surgical clips to be scored. In four external video sequences, a threshold calibrated on synthetic scenes falsely flags both static recordings. Together, these findings show how to interpret a negative screen in terms of the motions and amplitudes that the available views and the detector could reveal.
\end{abstract}
\begin{quote}\small\textbf{Keywords.} Structure from motion, Nonrigidity screening, Observability, Detectability, Reprojection residuals\end{quote}
\vspace{1em}
\end{@twocolumnfalse}}]

\section{Introduction}
Sparse structure from motion (SfM) estimates camera poses and three-dimensional structure under a static-scene assumption. SfM can still register most views of a deforming scene and produce plausible geometry because the camera and point estimates absorb part of the change. Registration success and visual plausibility therefore leave an important question unanswered: does the reconstructed static model adequately represent the observations collected over time?

We examine reprojection residuals: the differences between observed image points and the projections predicted by the reconstruction. A median or upper quantile summarizes their magnitude but discards which points and views contribute to the errors. Our question is whether the organization of these residuals across space and time provides additional evidence that the static-scene assumption is inadequate.

We propose a diagnostic that first removes residual components explainable by small corrections to the estimated points and cameras, together with simple spatial trends that change over time. It then measures how strongly the remaining errors follow predefined spatial and temporal patterns. The resulting score is compared with scores from static acquisitions; insufficient evidence or failed numerical checks lead to abstention. A positive screen indicates structured inconsistency in the supplied reconstruction. Calibration, correspondence and reconstruction errors can produce similar patterns. A negative screen is possible even when the scene changes, if the available views or the tested patterns are insensitive to that change.

A global reconstruction covers the full sequence, while a local reconstruction uses a short image window. Comparing overlapping local reconstructions with the global model can reveal changes obscured by fitting the whole sequence at once. The comparison requires matching points and aligning the extra reconstructions, and it depends on how the local cameras were estimated.

This article distinguishes three questions. First, can two different physical states yield the same sparse image tracks, and what extra observation separates them? Second, for a specified geometry, motion family and noise model, what response survives the fitted static model and how much of it reaches the score? Third, does the score's behavior persist as the reconstruction inputs, calibration and comparison methods change? We use the established ray ambiguity as a geometric baseline and derive conditional detection limits from the residual and modal gains. The evaluation combines synthetic, measured-shape and real-video evidence. The original tubular comparison uses four development-exposed families. Factorial ablations and external recordings test where the apparent benefit over simpler scores transfers. Synchronized-view tests distinguish the value of an additional view from a benefit of adaptive selection.

Online Resource~1 follows the article and contains the full methods, results and reproduction instructions. Online Resource~2 contains historical tables and protocols, and Online Resource~3 supplies code, protocols and selected results for the additional evaluations. See the \hyperref[sec:online_resources]{supplementary material descriptions} for the corresponding files.

\section{Related work}
\subsection{Residual structure and adjustment responses}
The proposed diagnostic connects bundle-adjustment sensitivity analysis with matched-subspace detection. Bundle adjustment supplies a local model of how point and camera perturbations change image predictions, including gauge freedom and sparse elimination~\cite{triggs}. Nuisance projection and matched-subspace detectors supply established ways to test residual alignment with a specified alternative, including unknown-noise formulations~\cite{scharf}. Geodetic deformation analysis likewise uses congruence tests, extended observation models and correlated adjustment information~\cite{velsink,velsink_extend}. Here these principles define which first-order reconstruction adjustments are removed before examining the remaining image residuals.

Flood et al. compress SfM reprojection information for low-memory map merging and discuss scene-change detection through changes in cost~\cite{flood}. This is a direct precedent for compressed SfM residual evidence. Our question concerns inconsistencies within one temporal reconstruction. We test a fixed quadratic spatial--temporal alternative after removing point, camera and affine-field responses, with empirical static calibration and explicit numerical abstentions. Our evaluation compares the stated readouts and separates physical ambiguity, nuisance removal, subspace sensitivity and empirical transport. These distinctions also explain why the magnitude of a reconstruction residual is an incomplete account of what a camera could have observed.

\subsection{Global--local geometric evidence}
Non-affine displacement analysis~\cite{falk} and spatial autocorrelation~\cite{moran} motivate removing broad geometric trends before examining coherent local differences. A spatial-coherence statistic combines robust affine removal, neighborhood averaging and a shuffled-amplitude reference. Its component controls test whether that combination adds value. A decision cascade selects among calibrated local diagnostics according to how the cameras were estimated, drawing on algorithm selection and selective classification~\cite{rice,elyaniv}.

Cloud-change methods provide additional comparisons. Multiscale model-to-model cloud comparison (M3C2) measures changes along locally estimated normals~\cite{lague}. Its error-propagation extension (M3C2-EP) accounts for measurement and registration uncertainty~\cite{m3c2ep}. The location and orientation united graph comparison (M3C2-LOG) combines both types of features~\cite{m3c2log}. We use native py4dgeo M3C2 distances~\cite{py4dgeo} and named lightweight graph-feature adaptations on the supplied matched pairs. These empirical sequence readouts have different inputs and uncertainty assumptions from the original topographic applications. We therefore report each adaptation and its usable support when interpreting the screening results.

\subsection{Deformable reconstruction}
Yunus et al.~\cite[Sec.~2.3.1]{yunus2024} review the depth ambiguity of monocular nonrigid reconstruction. Displacement along a viewing ray leaves the projected pixel unchanged, and moving the camera need not resolve the ambiguity when the scene changes between views. Nonrigid simultaneous localization and mapping (NR-SLAM) estimates cameras and nonrigid geometry~\cite{nrslam}. Local-EndoGS~\cite{localendogs} and the \href{https://arxiv.org/abs/2610.02343v1}{SCOPE-4D preprint} (Zhou et al., 2026) address dynamic surgical reconstruction. Our diagnostic uses estimated sparse geometry and its image observations to screen a whole acquisition for structured inconsistency. RoMo and GeoMotion estimate moving regions in color video~\cite{romo,geomotion}. We use them as video-only comparisons alongside epipolar fitting on shared tracks and depth-based geometry checks.

\section{Projected residual screening from a global SfM model}\label{sec:global}
\begin{figure*}[!htbp]\centering
\includegraphics[width=\linewidth]{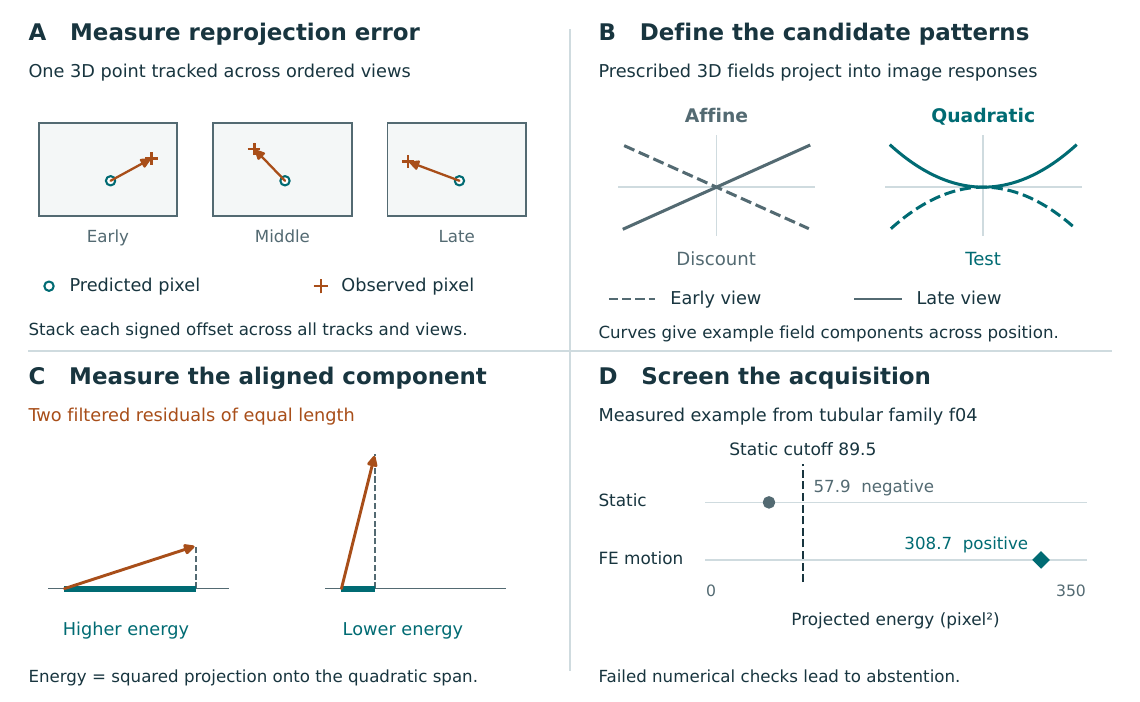}
\caption{Projected residual screening from one SfM model. Panel A shows arrows running from predicted to observed pixels for one track. In panel B, the early and late curves show one component of a prescribed field across position. In panel C, after filtering out point, camera and affine responses, two residuals of equal length make different angles with the surviving quadratic span (horizontal). The squared length of each teal projection gives its modal energy. In panel D, computed energies from the static and finite-element (FE)-derived acquisitions of family f04 are compared with the maximum static energy from the other three families. Panels A--C are schematic.}\label{fig:globalexplanation}
\end{figure*}

A static reconstruction may fit most image observations closely even when its reprojection errors vary systematically across points and views. We ask whether those errors retain a pattern that small adjustments to the estimated cameras and points cannot explain. We first remove the responses to such adjustments and to broad affine changes over time. We then measure how much of the remaining residual energy lies along prescribed quadratic patterns that vary across views. Comparing this energy with a cutoff calibrated on static acquisitions gives a screen for the whole sequence (Figure~\ref{fig:globalexplanation}). If the observations cannot support the calculation, the diagnostic abstains.

\subsection{Image residuals and adjustment responses}
A residual can arise because the supplied static reconstruction is imperfect. Before using its structure as evidence, we identify the image changes that small corrections to the reconstruction can explain. Let $\hat C$ and $\hat X$ be the estimated cameras and points, $y$ the observed pixel coordinates stacked into a vector, and $\pi$ the projection function. The residual and the point and camera response matrices (Jacobians) are
\begin{equation}
 r=y-\pi(\hat C,\hat X),\qquad A=\frac{\partial\pi}{\partial X},\quad B=\frac{\partial\pi}{\partial C}.
\end{equation}
The columns of $A,B$ describe changes in image predictions available to the linearized static model. We will remove their span from the residuals. To fix the arbitrary world frame and scale, seven gauge coordinates are removed: the first camera pose and a nonzero translation component of a second camera in the normalized chart.

The inputs come from the COLMAP model outputs~\cite{colmap}. The diagnostic uses the estimated structure, camera poses and image observations, with fixed pinhole intrinsics and track identities. A 3D point cloud alone lacks the required image observations and view order. The original comparison uses point refinement as an experimental preparation (Section~\ref{sec:protocol}).

\subsection{Spatial--temporal patterns to remove and test}
Having identified static adjustment responses, we need a specific alternative to test in the remaining errors. We construct prescribed 3D fields varying across points and views, then convert them into image changes that can be compared directly with $r$.

For point centroid $\bar X$, set $L=q_{.90}(\|\hat X_i-\bar X\|)$ and $p_i=(\hat X_i-\bar X)/L$; $q_{.90}$ is the linearly interpolated 90th percentile. The ten spatial features are
\begin{equation}
 \begin{aligned}
 b(p)={}&(1,p_x,p_y,p_z,p_x^2,p_y^2,p_z^2,\\
       &\sqrt2p_xp_y,\sqrt2p_xp_z,\sqrt2p_yp_z).
 \end{aligned}
\end{equation}
Views ordered by image name receive evenly spaced $\tau_j\in[-1,1]$. Combining each feature with $\tau_j,\tau_j^2$ and three world axes gives sixty fields. Field $(f,t,a)$ assigns observation $(i,j)$ the perturbation $L b_f(p_i)\tau_j^t e_a$, with unit world-axis vector $e_a$ and $t\in\{1,2\}$. The point-projection Jacobian converts each field into a pixel-response column. Dividing by its pixel root mean square (RMS) over observations gives $H$ with comparable column scales; the common extent factor cancels in this normalization.

We group the twenty-four columns for constant and linear spatial features into the affine nuisance matrix $H_a$. The thirty-six quadratic columns form the tested alternative $H_q$. This design can remove real affine deformation and retain structured reconstruction errors. The polynomial fields are prescribed independently of vibration or finite-element eigenmodes.

The evaluated 90-view acquisitions span one ramp/plateau/return cycle. Here $\tau,\tau^2$ supply fixed low-order contrasts, without reproducing that cycle exactly or modelling physical dynamics. Constant point shifts already lie in the adjustment space. Fast, localized, discontinuous or higher-order motion, and motion removed with nuisance responses, can be missed; the basis is not claimed optimal.

\subsection{Projected energy and the screening decision}
A quadratic pattern already explained by the discounted responses cannot supply additional evidence. We therefore apply the same removal to the measured residuals and the candidate fields. Let $N=[A,B,H_a]$ collect the discounted directions and $P_N^\perp$ project off its numerically identifiable span. Filtering $r,H_q$ and choosing an orthonormal basis $Q_q$ for the surviving quadratic space gives
\begin{align}
 e_0&=P_N^\perp r,& Q_q&=\operatorname{orth}(P_N^\perp H_q),\\
 \Delta_{\rm mod}&=\|Q_q^{\mathsf T}e_0\|^2,&
 e_1&=e_0-Q_qQ_q^{\mathsf T}e_0.
\end{align}
As Figure~\ref{fig:globalexplanation}C illustrates, equally long filtered residuals can have different energies because their projections have different lengths. Here $e_0$ is the filtered residual, $\Delta_{\rm mod}$ its energy along $Q_q$, and $e_1$ the error left after removing that component too. The energy is our primary acquisition score; comparison with static acquisitions determines the screen (Section~\ref{sec:decision}).

Numerical ranks account for overlapping directions and observation geometry. Across twelve primary models, the 24 affine columns retain rank ten after point/camera elimination, while the quadratic alternative retains rank thirty-six. Energy sums squared pixels and can depend on observation count, track multiplicity, visibility, resolution and residual scale. We also report an optional normalization by the remaining residual variance. With $q=\operatorname{rank}(P_N^\perp H_q)$ and $\nu=2n_{\rm obs}-\operatorname{rank}(N)-q$ for $n_{\rm obs}$ image observations,
\begin{equation}
 F_{\rm mod}=\frac{\Delta_{\rm mod}/q}{\|e_1\|^2/\nu}.
\end{equation}
Both readouts use the same operator with their own static calibration; they give identical decisions in the original estimator-polished comparison. Estimated geometry, correlated observations and selected tracks do not justify a nominal $F$ null law, $p$-values or physical probabilities. To assess what pattern alignment adds, magnitude controls use median and 95th-percentile (q95) raw residual norms and the degrees-of-freedom-adjusted RMS of $e_0$ (Supplement S1.1).

\subsection{Numerical availability}
The original score is reported only when the observations support the projected spaces and remaining residual variance. Rank, conditioning and null-space checks establish this availability; an invalid point-information block makes the whole model abstain, without dropping tracks. Matched magnitude controls use that same rule, although an unrestricted median could still be computed. An abstention therefore remains distinct from a negative screen. Bundle-adjustment block elimination avoids a dense observation-by-parameter matrix; Supplement S1 details the factorization, gauge and numerical thresholds.

\section{What the observations and the score can resolve}\label{sec:observability}

\begin{figure*}[!htbp]\centering
\includegraphics[width=\linewidth]{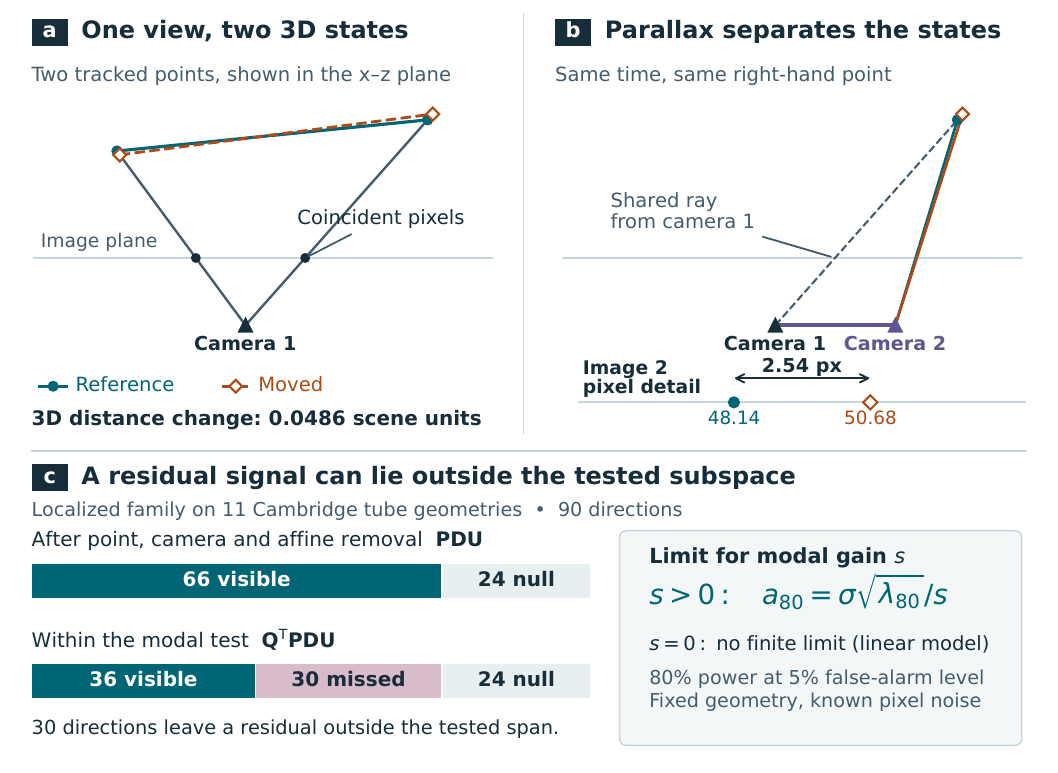}
\caption{Observability and detection limits. (a) Two points from the perspective fixture viewed in the $x$--$z$ plane. Their 3D separation changes by 0.0486 scene units, yet each point stays on its original ray and projects to the same pixel in camera 1. (b) A simultaneous camera displaced by 0.30 scene units sees the right-hand point at two different horizontal pixel coordinates. Its rays distinguish the states. (c) The localized motion family has 90 dimensions on each of the 11 Cambridge tube geometries. After removing point, camera and affine responses, 66 directions remain observable to first order. The modal test retains 36, with a further 30 directions contributing to the residual but yielding no modal response. The bars report dimensions under the relative numerical rank tolerance of $10^{-9}$. At fixed geometry under known independent Gaussian pixel noise, the RMS amplitude $a_{80}$ needed for 80\% power at a 5\% false-alarm level depends on the direction's modal gain $s$}
\label{fig:observability}
\end{figure*}

A physical change may leave the observed tracks unchanged. When the tracks do change, part of the response can be explained by the fitted static model or by the discounted affine fields. The detector tests a subspace of the remaining signal (Figure~\ref{fig:observability}).

\subsection{Viewing-ray ambiguity in sparse tracks}
Let $\mathbf C_j$ be the centre of camera $j$ and $\mathbf X_i$ a static point.
For any positive scalar $\gamma_{ij}$, define the moving alternative
\begin{equation}
 \mathbf X'_{ij}=\mathbf C_j+\gamma_{ij}(\mathbf X_i-\mathbf C_j).
 \label{eq:ray}
\end{equation}
In camera coordinates, the two points differ by a positive scale. Perspective division therefore gives the same pixel for both, even if the camera pose is known. The scalars can vary across points and time so that interpoint distances change. With the same point identities and visibility, any rule whose sole observations are these tracks receives identical inputs in the static and moving cases. An additional camera observing the \emph{same state} distinguishes a displaced point if its centre is off the first viewing ray and it observes the corresponding point. A later view can see a point after it returns to its static position. This argument concerns sparse correspondences; visibility, appearance and physical constraints can supply additional information.

\subsection{Conditional sensitivity to a prescribed motion family}
To distinguish what survives adjustment from what the score actually tests, let $\mathbf D$ map small 3D displacements of the visible points to pixels at a fixed geometry, and let $\mathbf U$ span a prescribed family of displacements. Its columns are orthonormal in root-mean-square displacement over the represented point--view pairs. Write $\mathbf P=P_N^\perp$ for nuisance removal and $\mathbf Q=Q_q$ for the score's tested space. The singular gains of $\mathbf P\mathbf D\mathbf U$ describe sensitivity remaining after the static adjustments and discounted affine fields. The gains of $\mathbf Q^{\mathsf T}\mathbf P\mathbf D\mathbf U$ describe the part captured by modal energy. A direction with zero gain in the second map can have a nonzero response in the first.

At fixed geometry, assume independent zero-mean Gaussian pixel errors of known variance $\sigma^2$ and zero projected mean in the static case. Under the linear response model, a direction with modal gain $s$ and RMS amplitude $a$ gives a noncentral chi-square law for $\Delta_{\rm mod}/\sigma^2$. Its degrees of freedom are $q$ and its first-order noncentrality is $(as/\sigma)^2$. Let $\lambda_{80}$ be the noncentrality giving 80\% power at the central-law 5\% cutoff. Then
\begin{equation}
 a_{80}=\sigma\sqrt{\lambda_{80}}/s \qquad (s>0).
 \label{eq:limit}
\end{equation}
A zero-gain direction has no finite limit in this linear model. The family, noise level, observation support and scale are inputs to this calculation. Equations~\eqref{eq:ray} and \eqref{eq:limit} concern different limits. The former gives indistinguishable full projections. The latter characterizes a particular local score and its alternative. The empirical thresholds used on estimated reconstructions remain those of Section~\ref{sec:decision}.

\subsection{Observable track directions and focal scale}
A rank-deficient track can still constrain the images. If $\mathbf U_i$ contains the retained left singular vectors of its point Jacobian, the least-squares elimination of that point projects all rows for the track with $\mathbf I-\mathbf U_i\mathbf U_i^{\mathsf T}$. An orthonormal complement retains $2n_i-r_i$ coordinates for $n_i$ pixel observations and rank $r_i$. Eliminating these blocks before the shared camera and affine responses preserves the joint least-squares objective for the retained spaces. The rank-aware variant keeps the pixels without estimating unobserved depth and still requires usable camera, mode and residual information.

At fixed pinhole geometry and principal points, a common positive focal multiplier $s_f$ gives the exact residual $\mathbf y-\mathbf c-s_f\mathbf v$. The point, camera and field Jacobians all scale by $s_f$, leaving their spans unchanged. Adding $\mathbf v$ to the discounted directions makes the projected residual and alternative space invariant to $s_f$ in exact arithmetic at fixed ranks. Refitting the geometry, using independent focal changes, or dividing by the misspecified focal length can change the resulting readouts. The variants and numerical checks are specified in Online Resource~1, S12.

\section{Global--local comparisons}\label{sec:local}
\begin{figure*}[!htbp]\centering
\includegraphics[width=\linewidth]{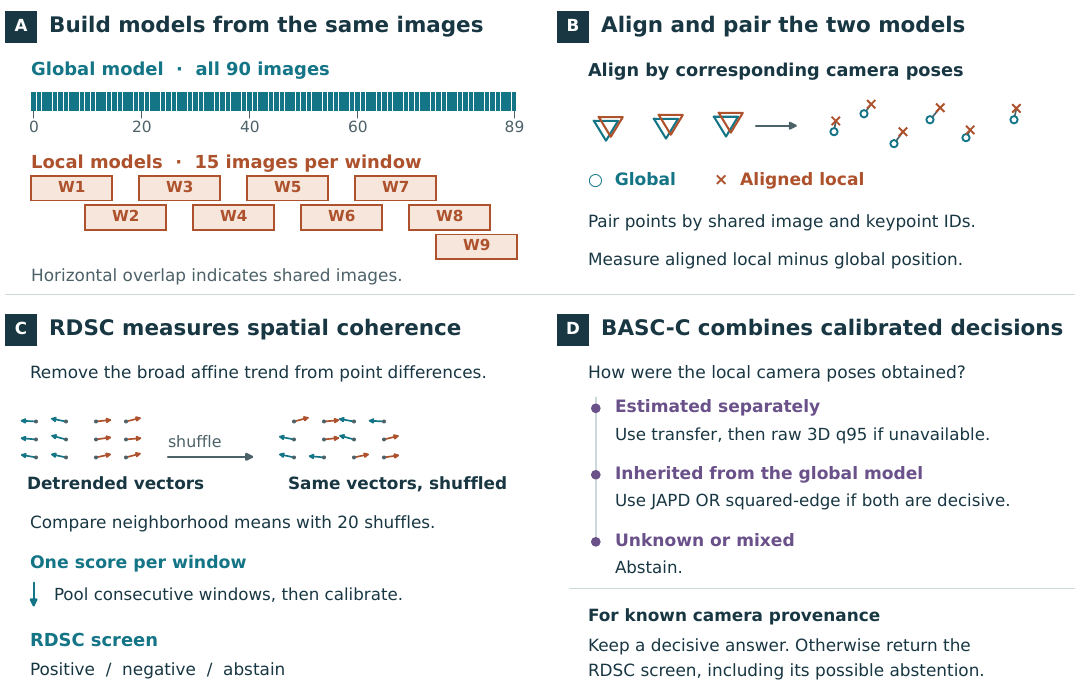}
\caption{From shared images to global--local screening. In panel A, each tick represents one of the 90 images used by the global reconstruction. The nine bars mark overlapping 15-image bundles on the same time axis. In panel B, camera poses determine the alignment used to compare corresponding points. Panel C shows how RDSC compares local agreement among residual vectors with a shuffled reference. Panel D shows how BASC-C uses the origin of the local camera poses to choose among calibrated expert decisions, including joint adjustment-propagated disagreement (JAPD). The point pairs and arrows are schematic.}\label{fig:localworkflow}
\end{figure*}

Fitting one geometry to the full sequence can obscure changes that are easier to see over shorter intervals. In the Bundles workflow~\cite{thesis}, reconstructions of overlapping image windows provide local models for comparison with the full sequence. We align the models using corresponding camera poses and then compare matched points. Robust detrended spatial coherence (RDSC) measures whether nearby points disagree in similar directions and pools that evidence across windows. Bundles applicability-aware selective composition with cascade preservation (BASC-C) uses the origin of the local camera poses to choose among calibrated diagnostic decisions. These two uses of short-window reconstructions (Figure~\ref{fig:localworkflow}) provide acquisition-level decisions that we can compare with the single-model screen.

\subsection{Aligned differences and spatial coherence}
We pair global/local points through shared image/keypoint identities. A proper similarity estimated from corresponding camera poses aligns the models; it is never fitted to the point differences subsequently tested. Let $d_i$ denote aligned local-minus-global differences and $e_i$ their residuals after robust affine removal (Supplement S3; Figure S2).

RDSC measures how much spatially coherent disagreement remains after affine detrending. Averaging neighboring residual vectors retains components that point in similar directions. Shuffling the same vectors gives a reference for the agreement expected without their original spatial arrangement. The score combines the observed amplitude $a$ with its ratio to the shuffled reference $b$:
\begin{equation}
 \begin{aligned}
 a&=\frac{q_{.95}(\|\,|\mathcal N_i|^{-1}\sum_{j\in\mathcal N_i}e_j\|)}{L},\\
 b&=\operatorname{median}_{h=1}^{20}a(e_{\sigma_h}),\\
 R&=\frac{a^2}{\max(b,\eta)}.
 \end{aligned}
\end{equation}
Here $L$ is the matched global extent, $\mathcal N_i$ normally contains the query and eight neighbors, and $\eta$ is the smallest positive normal float64 value. Each permutation $\sigma_h$ uses the same graph and pooling. Controls $a$, $a/b$, additive contrasts and raw quantiles assess the combined score. Affine removal can suppress real motion, and small $b$ can amplify weak signals. Supplement S3.2 details the fit, permutations and tie handling, and Figure S3 illustrates the stages. Temporal pooling and static calibration then turn these window scores into an acquisition decision (Section~\ref{sec:decision}).

\subsection{Combining calibrated expert decisions}
BASC-C combines expert acquisition-level decisions: positive, negative or unavailable. Its choice of a preferred answer depends on how local cameras were obtained: separate camera estimates permit a transfer contrast, while inherited cameras provide no separate camera-estimation evidence. Shared observations still couple separately estimated models, so this rule concerns applicability rather than accuracy.

The transfer expert compares middle-view projection errors after endpoint triangulation under each camera set. Raw 3D q95 instead summarizes aligned point differences. Joint adjustment-propagated disagreement (JAPD) evaluates those differences relative to estimated point/camera sensitivity to shared image-observation perturbations. Squared-edge contrast applies joint-response information to neighboring squared lengths. These conditional measures are not calibrated physical-deformation probabilities.

For separate cameras, BASC-C uses transfer, then raw 3D q95 if unavailable. For inherited cameras, both JAPD and squared-edge must return a decision. The preferred answer is positive if either expert screens positive. The cascade preserves a decisive preferred answer and otherwise uses RDSC. The cascade abstains when camera provenance is unknown or mixed. Supplement S3--S4 defines the experts, interval pooling and fallback-removal controls; Figure S4 shows the routes.

\section{Experimental protocol}\label{sec:protocol}
For the original comparisons, each requested acquisition receives a positive or negative screen, or an abstention. Temporal pooling and static calibration turn the retained scores into sequence decisions. Section~\ref{sec:campaign} explains the added experiments. It reports conditional power, transfer of scores and cutoffs, and geometry error under their respective observation and reference assumptions.

\subsection{Comparison questions and supplied inputs}
Table~\ref{tab:cohorts} shows which acquisitions address each of the four research questions (RQ1--RQ4) and where their inputs overlap.
\begin{description}
\setlength{\itemsep}{0pt}
\item[RQ1.] Does the spatial and temporal structure of reprojection errors reveal moving acquisitions that residual-magnitude summaries miss?
\item[RQ2.] Which detections persist across changes in appearance and reconstruction pipeline, and do decisions change when we use the original point coordinates?
\item[RQ3.] What screening evidence do short-window reconstructions add, and how do the local methods compare with their simpler components?
\item[RQ4.] How does static calibration affect the decisions, what reconstruction coverage is needed for a screen, and what does scoring cost once the models exist?
\end{description}

\begin{table*}[!htbp]\centering\small
\begin{tabularx}{\linewidth}{>{\raggedright\arraybackslash}p{.24\linewidth}>{\raggedright\arraybackslash}X>{\raggedright\arraybackslash}X}\toprule
Cohort / check & Available evidence & Purpose and dependence\\\midrule
Primary comparison & 12 acquisitions: 8 positive, 4 static; f01--f04 & RQ1/RQ3; four development-exposed families\\
Appearance/pipeline & 20 positive exports; f01/f04 & RQ2; reused geometry, no matched negatives\\
Original exports & Same 32 global models without point refinement & RQ2; input comparison, no new acquisitions\\
Matched local models & 216 window/regime records from the 12 primary acquisitions & RQ3; overlapping windows and shared images\\
Generated utility & 26,400 specified point-pair problems & RQ3; tests RDSC complexity, not scene generalization\\
Historical/physical & Retained model collections and source-specific labels & RQ4; dependent inputs and coverage limits\\
Colonoscopy 3D Video Dataset v2 (C3VDv2) & 9 requested actuation recordings; 1 lacks raw images & RQ4; insufficient SfM coverage, not discrimination\\\bottomrule
\end{tabularx}
\caption{Evaluation map. Rows describe different questions and reuse inputs; their counts must not be added as independent samples. Detailed populations, support and unavailable requests are in Supplement S2 and S6--S7.}\label{tab:cohorts}
\end{table*}

\begin{figure*}[!htbp]\centering
\includegraphics[width=.90\linewidth]{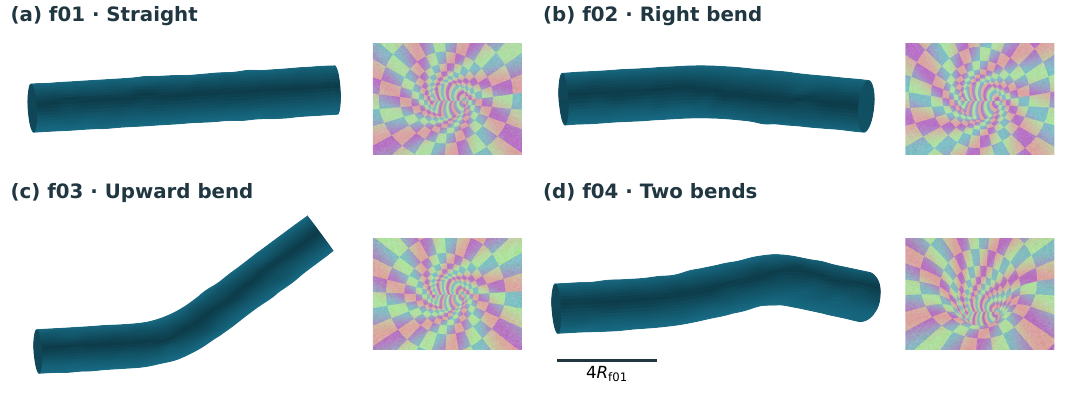}
\caption{Primary evaluation families and retained views. (a--d) f01 straight, f02 right bend, f03 upward bend and f04 two bends: source wall (left), color (red, green and blue, RGB) image at frame 52 of the static-scene acquisition with a moving camera (right). Meshes share an orthographic projection and generating-coordinate scale normalized by f01's reference radius $R_{\mathrm{f01}}$. Identical bounds preserve aspect ratio and scale; the shared bar is $4R_{\mathrm{f01}}$, although other families can have different radii. RGB uses a common square-root transform. These acquisition illustrations are excluded from diagnostic inference.}\label{fig:atlas}
\end{figure*}

Each of the four tubular families has one static acquisition, one with finite-element (FE)-derived motion and one with separately prescribed motion without FE. The cohort contains eight positives and four statics, with 729,539 reconstructed points in total. FE-derived motion uses finite-element bases with prescribed timing, rather than physiological dynamics (Figures~\ref{fig:atlas} and S1). Each global model spans ninety images; local models use nine overlapping fifteen-image windows starting at $0,10,\ldots,70,75$, with cameras estimated separately or inherited from the global model.

To match estimator states before comparing global and local evidence, the \emph{estimator-polished comparison} (EPC) optimizes only 3D points to reduce reprojection error, starting from their supplied coordinates. Cameras, intrinsics, tracks and observed pixels stay fixed; Supplement S2.1 specifies the deterministic objective, initialization, stopping and failure rules. Original-export replay then tests the three-text-file interface without refinement arrays, recalibrating from original global scores rather than assuming score invariance.

The original transfer test uses cutoffs fixed from the primary static acquisitions and evaluates twenty global exports from two primary families, five appearances and two reconstruction pipelines. Pipeline A uses the colonoscopy-specific COLMAP fork~\cite{colmap_colonoscopy}. Pipeline B uses the fork of upstream COLMAP 4.2.0 with mapper-ordering corrections~\cite{colmap_ordering}. All exports span the prescribed motion cycle. The design measures retention of positive detections across the combined changes. The twenty unique exports are the counting unit (Supplement S2.2).

\subsection{Experimental inputs and cited resource versions}\label{sec:provenance}
The models originate in benchmark scientific release 1.0.0 and its reproduction material~\cite{benchmark,benchmark_support}. The diagnostic input deposit provides the exact model exports used here~\cite{screeninginputs}. The \href{https://arxiv.org/abs/\ResourcePreprintID}{companion benchmark preprint} (Cirauqui-Viloria, 2026) describes the acquisition protocol. LumenMorph v1.0.2~\cite{lumenmorph} is the historical software reference. The expanded benchmark and support deposits~\cite{benchmark_current,benchmark_support_current} are available as collection 2.0.3 (\texttt{v2.0.3-docs1}), with LumenMorph v2.0.3~\cite{lumenmorph_current}. The reported benchmark comparisons use the original set of SfM models.

\subsection{Sequence decisions, calibration and coverage}\label{sec:decision}
A global readout already describes one acquisition. Local experts first require temporal pooling: at least eight of nine requested windows must be usable, with a consecutive run of three for transfer or two for other experts. Within each eligible run we take the minimum; the sequence score is the maximum of those run minima. This requires scores to persist across adjacent windows rather than depend on an isolated peak. For an interval-valued score, we pool both endpoints separately, conditional on the available windows and frontends (Supplement S4). Missing indices are never joined.

Each readout in the primary comparison and its original transfer test uses the maximum primary static score from families other than the target's. The later factorial transfer test uses all four primary statics (Supplement S12). A valid scalar score is positive if it exceeds its cutoff and negative otherwise. BASC-C combines already calibrated expert decisions. Failed numerical checks, insufficient coverage or unresolved target/calibration intervals cause abstention (Supplement S4, including distinct historical schedules). Every requested acquisition stays in the denominator; abstentions are counted separately from errors.

The calibration rule also determines how static outcomes must be read. With four statics and a unique maximum, the largest static score exceeds the maximum of the other three, necessarily producing one static false alarm. True-negative (TN) and false-positive (FP) counts therefore describe this procedure rather than population specificity. Positive detections compare methods under this rule, without establishing a universal ranking under other threshold distributions. Full and delete-one calibration definitions are retained; folds and overlapping windows are not extra acquisitions.

All primary/transfer exports register ninety $640\times480$ images. Supplement S2 gives each request's observations, points, track multiplicity, ranks, residual degrees of freedom and score units. Since modal energy sums squared projected pixel residuals, its unchanged cutoffs are tested only on these support distributions. Portability across resolutions, track densities or observation-noise distributions remains unestablished for both energy and variance normalization.

\subsection{Comparisons, numerical verification and cost}
Magnitude controls test what pattern alignment adds: median/q95 reprojection norm and nuisance-null RMS. Matched local comparisons then use RDSC, BASC-C, elementary components and a native py4dgeo 1.2.0 M3C2 distance wrapper on the same supplied pairs with equal static-calibration access, while their inference requirements differ. M3C2-support raw controls use its usable points to separate support selection from distance design.

The available inputs also limit what can be compared. LOG-lite and its zero-distance (ZD) extension adapt location/orientation features; they do not reproduce complete M3C2-LOG. Their failures and outcomes remain in Supplement S5--S6. M3C2-EP requires sensor/scan and propagated registration uncertainty not qualified for these sparse exports, so receives no benchmark score. Supplement S5 lists other unexecuted systems and their input/source limits. Reconstruction systems with different outputs remain related work.

Before interpreting scores, direct least-squares/\allowbreak streamed comparisons and independent actual-model matrix audits check numerical implementation within fixed tolerances (Supplement S2.3). Profiling separately measures central processing unit (CPU) time and peak resident set size (RSS) on all twelve primary acquisitions, using fresh processes, single-thread libraries and a pinned Linux runtime; repeats check output agreement. Loading/scoring and wall time are in Supplement S8. Timing starts after SfM: local models already exist, and BASC-C computes all evidence rather than an optimized lazy cascade.

Scripts render the figures from data and explicit drawing instructions. OpenAI Codex assisted with manuscript editing and with developing and checking the code used for the analysis and figures. The author takes responsibility for the final work.

\section{Observability and transfer on additional evidence}\label{sec:campaign}

\subsection{Which changes survive the projection?}
In the exact perspective fixture, one interpoint distance changes by 0.0486 scene units while corresponding image differences over twelve primary views stay below $5.7\times10^{-14}$ pixels. A displaced camera observing the same state produces a difference of 3.663 pixels at the most affected point. An independent rational-arithmetic check also verifies the equality of the primary projections. This establishes an ambiguity for track-based inference and the role of a simultaneous second ray.

For the conditional power check we used the unloaded marker geometries of eleven Cambridge tube configurations~\cite{cambridge_tube}. We projected smooth quadratic, localized and rapid displacements through specified pinhole cameras. The simulations add independent Gaussian noise with a standard deviation of 0.3 or 1 pixel while holding geometry and subspaces fixed. The smooth-quadratic family retained 36 directions in the modal test. The localized family spans 90 directions. After removing point, camera and affine responses, the residual map has rank 66. The modal projection retains rank 36 and misses 30 additional directions that still affect the residual. The empirical false-alarm fraction was 5.22\% in the simulated Gaussian check at a nominal 5\%. For 369 in-range direction, amplitude and noise combinations, the largest difference between predicted linear power and power computed with finite perspective displacement was 0.104. Another 225 combinations required amplitudes outside the specified range. These counts are numerical checks on measured geometries with projected coordinates, not independent validation of power after reconstructing from real video.

In a separate check, we generated synthetic pixel coordinates from measured rest and deformed shapes of the same Cambridge configurations. Camera poses and a static structure were refit with privileged intrinsics, correspondences and an unloaded seed. Fifty-seven of 66 fit requests converged. At 0.3-pixel noise, the weak and full motion arms gave 10/11 and 8/11 positives for modal energy and each magnitude control. At 1 pixel, the corresponding counts were 3/11 and 1/11, with calibration available for only three requests in each arm. The modal and magnitude readouts give the same counts at both noise levels.

\subsection{Attribution, alternative readouts and score transport}
The eight-arm factorial independently switches point, camera and affine elimination on the 32 original exports. Each variant is calibrated on the same statics using its own scores. The complete arm reproduces the canonical energy in all 31 scored exports. Two simpler arms, point plus camera and point plus affine, each match its 19/20 transfer count and its one numerical abstention. On the same 32 exports, camera-conditioned Sampson q95 produces 8/8 core and 13/20 transfer positives. Sequential MAGSAC and the tested Progressive-X configuration~\cite{magsac,progressivex} give zero secondary-motion sequence scores with the declared aggregation. One Progressive-X sampler also fails to fit any pair. These geometry-based comparators test different signatures. The results apply to the configurations and aggregation rules used here. Online Resource~1, S12 reports all factorial counts and refusals.

The rank-aware variant matches direct dense projection on mixed-rank fixtures and the canonical energy in the 31 originally scored exports. The last export still abstains because its remaining residual variance is numerically zero. Among seven requested StereoMIS clips~\cite{stereomis}, the rank-aware variant scores two fully registered clips that the original point-rank check rejected. Five event-positive clips from three subjects score above the 89.50 pixel$^2$ original-export cutoff. One request lacks camera calibration and one registers four of ninety views. Matched static sequences and independent tissue-displacement measurements are absent.

Holding the original cutoff fixed while halving pixel units causes one of eight positives to fall below it. Doubling pixel units incorrectly flags all four statics. Studentization and angular normalization preserve the original decisions under both unit changes. After 36 bounded point-and-camera refits with a shared focal perturbation of $\pm2\%$ or the native scale, removing the focal response reduces four and three static false alarms to one in the low and high arms, respectively. The absolute corrected gain can still drift after refitting. The largest drift occurs in a fit that stops at its iteration limit. Thus focal invariance at fixed geometry and threshold transfer after refitting are distinct.

\subsection{Real-video tests and geometry checks}
The fixed original-export cutoff detects the four moving Bonn sequences, flags the one scored static and abstains on the second static. In the four TUM Freiburg~3 recordings from a separate laboratory it detects both walking sequences and flags both statics. In TUM, pixel median and q95 rank both walking sequences above both statics. One static exceeds both under raw modal energy. When the recorded depth reference defines a two-of-four review budget, each of these magnitude summaries picks both higher-error reconstructions, while modal energy picks one. The reference evaluates registered tracks after estimating metric scale from the first ten depth-qualified views. Bonn/TUM labels describe scene motion rather than measured tissue deformation~\cite{bonn,tum}.

RoMo and GeoMotion~\cite{romo,geomotion} are RGB-only comparisons on the same six videos, using a fixed 16-frame schedule shared with the corresponding SfM runs. RoMo's 90-frame trial exceeds its time budget, and its reduced-frame run produces two static fallbacks with no trusted frames. The raw GeoMotion checkpoint scores all six videos, but only one moving scene exceeds both static scenes by median moving fraction. The reduced-frame modal screen scores four of six requests. These outputs measure different properties, and the comparison records eligibility, ranking and time separately from any thresholded claim.

DRENDS and D4D provide independent optical depth or geometry~\cite{drends,d4d}. The stereo depth estimates for both DRENDS sequences fall short of the specified accuracy and coverage criteria. In D4D, the monocular diagnostic abstains on all six requested clips from one specimen. Independent structured-light endpoint comparisons support the evaluation of stereo depth, but the released stereo depth change beats zero change in only one of six clips. Optical captures are offset from the paired RGB endpoints by 2.66--8.63 seconds, and their camera registration is privileged. These studies characterize physical input and evaluation limitations.

\subsection{What an additional simultaneous view supplies}
Across the eight controlled camera-selection regimes, the oracle improves on the retrospectively best fixed action by at most 8/720 displaced requests. In SceneHub4D~\cite{scenehub}, the replay compares two recorded, synchronized camera alternatives at 38 instants in four recordings. The selector chooses before reading the candidate image or independent anchor depth. In the changing kitchen region, 106 of 720 anchor feature requests meet the 10-cm tolerance under the adaptive rule, versus zero for maximum camera separation and 137 for the best fixed view. That fixed view attains the oracle count in each recording. This is evidence of geometric value from a suitable simultaneous view under the tested support, with no advantage over the best fixed action. The physical Cambridge rig further limits the disjoint reference design to at most four common triangulable markers per pressure state. Online Resource~1, S14 reports the capture designs and results.

\section{Results on the original benchmark}
We first compare primary screening decisions, then ask whether they persist under the specified input changes. The local-model comparisons examine additional detections, error patterns and the value of more elaborate scores. Finally, cost and reconstruction availability establish the conditions under which these diagnostics can be used. Errors and abstentions remain visible throughout.

\subsection{RQ1: residual structure versus magnitude}
The primary comparison tests the complete projected score against magnitude summaries under the same calibration rule. Modal energy detects all eight positives, versus five for pixel median and six for q95 or nuisance-null RMS (Table~\ref{tab:core}). Each readout correctly classifies three static acquisitions, flags one and scores all twelve requests. Thus the full operator recovers positives missed by magnitude summaries; Section~\ref{sec:campaign} tests the attribution to nuisance-removal stages on the original exports. Optional $F_{\rm mod}$ reproduces the energy decisions, with no additional observed benefit.

\begin{figure*}[!htbp]\centering
\includegraphics[width=\linewidth]{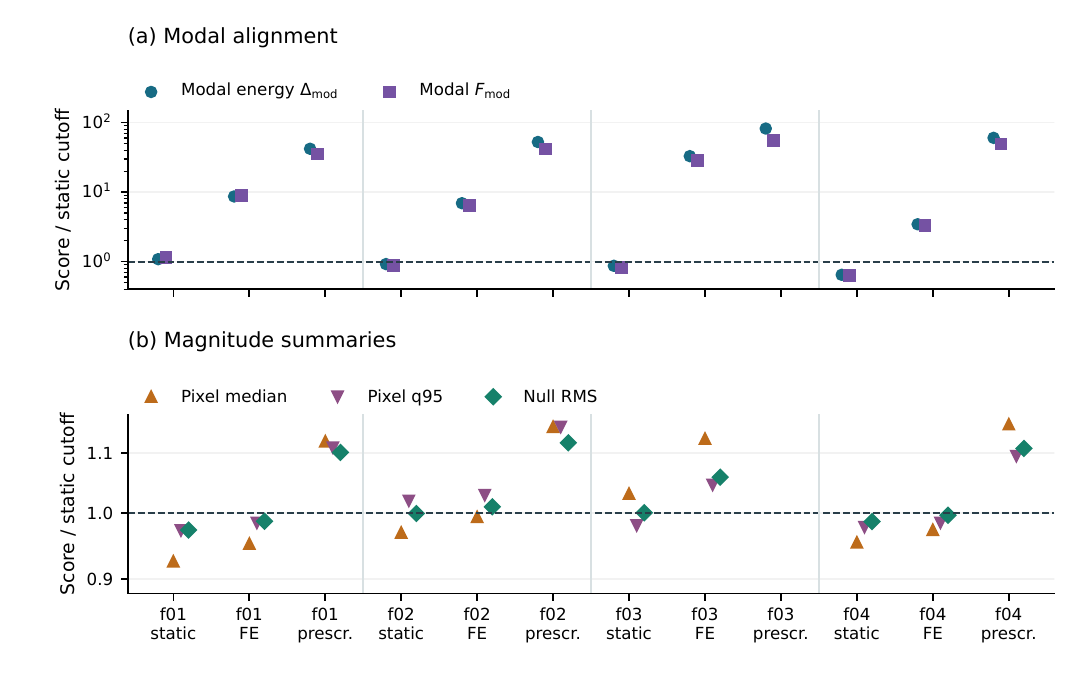}
\caption{Complete global-readout comparison. Scores are divided by each method's family-excluded static cutoff; values above one screen positive. (a) Modal energy and its studentized readout; (b) residual-magnitude summaries, shown on an expanded scale near the cutoff. The panels intentionally use different vertical scales. FE and prescr. identify the two prescribed-motion conditions. Every acquisition is shown. Energy and $F$ make identical decisions; the single static false positive follows the maximum-calibration design.}\label{fig:core}\end{figure*}

The static false alarm coexists with complete score separation: the lowest positive modal energy is 308.69 pixel$^2$, above the highest static energy of 89.53 pixel$^2$. The false alarm follows the family-excluded calibration rule (Section~\ref{sec:decision}), rather than positive/static overlap in this cohort. This descriptive gap establishes no margin on new geometries. Figure~\ref{fig:core} shows every acquisition against its family-specific cutoff.

\begin{table*}[!htbp]\centering\small\setlength{\tabcolsep}{4pt}
\begin{tabular}{lcc}\toprule
Readout & Core: TP/TN/FP/FN/A & Transfer: TP/FN/A\\\midrule
Modal energy $\Delta_{\rm mod}$ & 8/3/1/0/0 & 19/0/1 \\
Modal $F_{\rm mod}$ & 8/3/1/0/0 & 19/0/1 \\
Pixel median & 5/3/1/3/0 & 3/16/1 \\
Pixel q95 & 6/3/1/2/0 & 8/11/1 \\
Nuisance-null RMS & 6/3/1/2/0 & 5/14/1 \\
\bottomrule\end{tabular}

\caption{Projected residual readouts and magnitude controls. Primary cohort: twelve EPC acquisitions, eight positive/four static. Transfer: twenty positive-only exports with unchanged thresholds. TP/TN/FP/FN/A denote true positives, true negatives, false positives, false negatives and abstentions. Counts include every requested acquisition; global answers are counted once. Energy and $F_{\rm mod}$ are readouts of the same operator.}\label{tab:core}
\end{table*}

\subsection{RQ2: appearance and pipeline variations}
\begin{figure*}[!htbp]\centering
\includegraphics[width=\linewidth]{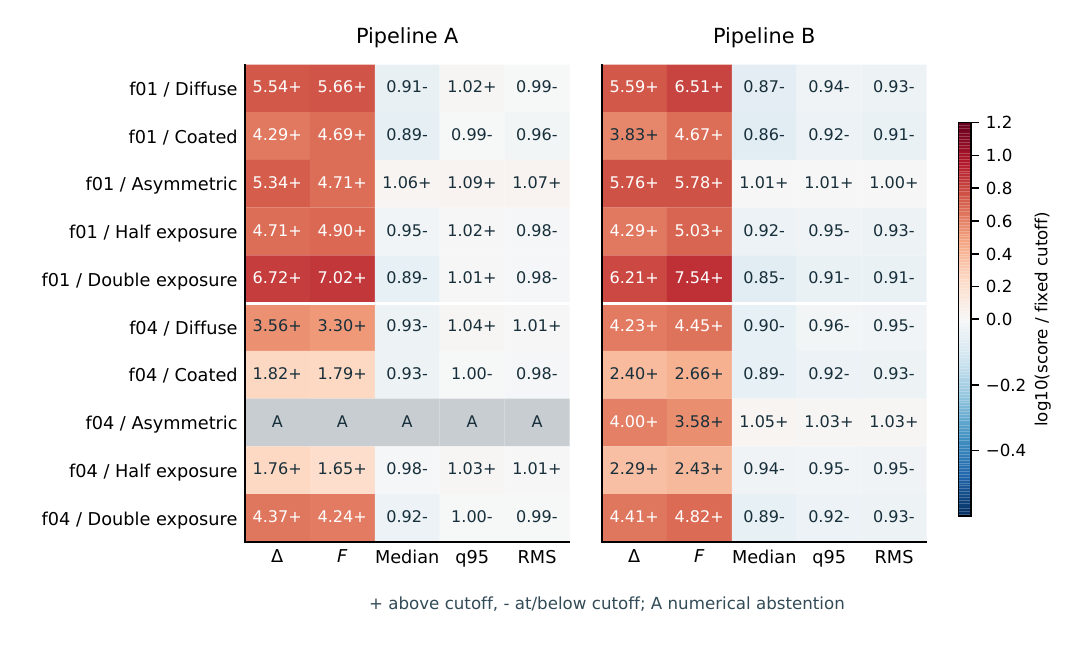}
\caption{Complete positive-only appearance/pipeline transfer. Cells show score divided by the static cutoff for the target's family. Color shows the base-10 logarithm of this ratio. Values are rounded to two decimals, while appended $+$ or $-$ signs use the unrounded decision. Gray A denotes the retained point-information-rank abstention. All twenty unique exports are shown; the two pipelines and five appearances reuse two core families and contain no matched appearance negatives.}\label{fig:transfer}\end{figure*}

We next test detection retention using the family-excluded cutoffs from the primary static acquisitions. Modal energy detects nineteen of twenty requested positive exports and abstains once, with zero false negatives among scored requests. The median, q95 and null RMS detect three, eight and five positives, respectively, and all abstain on the same export (Table~\ref{tab:core}). Supplement S12 explains the eligibility and calibration differences in the factorial comparison. Figure~\ref{fig:transfer} shows all variants, with decision signs taken before score rounding. Positive detections therefore largely persist across the specified changes. The variants reuse two families and lack matched negatives (Section~\ref{sec:protocol}), so this is a retention result rather than independent geometry or false-positive validation.

\subsection{RQ2: original exports and numerical availability}\label{sec:plain}
The next check asks whether the diagnostic requires EPC point refinement. Under the original eligibility and calibration rules, using only original COLMAP exports and recalculating the cutoffs preserves all 340 distinct decisions: 240 primary-cohort checks and 100 transfer checks. These are readout/calibration evaluations of thirty-two models, not 340 acquisitions; only ninety-six global text files are accessible. Ranks, statuses and failure reasons also agree. Transfer scores are bitwise equal; primary scores change by up to 0.47926 across differently scaled readouts. On these inputs, the three-file interface therefore reproduces decisions without further point optimization (Supplement S2.3).

The shared transfer abstention explains the numerical availability limit. In f04/\allowbreak asymmetric/\allowbreak pipeline A, all ninety images and 29,434 requested points are present, but one point-information block fails the rank condition. Its full support remains counted; no points are dropped or numerical floors lowered. With nineteen scored detections, both modal readouts pass the predefined requirement of at least eighteen scored requests and sixteen detections. The separate requirement that studentization improve on energy fails (Supplement S2.4).

\subsection{RQ3: added value of global--local evidence}
We now ask what the extra local inputs add, and whether compound scores outperform their components. With separate cameras, BASC-C, raw 3D median, JAPD and transfer match the global counts (Table~\ref{tab:published}). Coherent amplitude matches RDSC in both regimes (seven/six TP); ratio detects five in each. With inherited cameras, BASC-C adds a false alarm relative to JAPD because a decisive preferred error prevents fallback. Figure S5 and Supplement S6 retain every paired decision and all graph-feature variants.

\begin{table*}[!htbp]\centering\small\setlength{\tabcolsep}{4pt}
\begin{tabular}{lcc}\toprule
Readout & Independent cameras & Inherited cameras\\\midrule
\multicolumn{3}{l}{\emph{Global--local strategies}}\\
RDSC & 7/3/1/1/0 & 6/3/1/2/0 \\
BASC-C & 8/3/1/0/0 & 7/2/2/1/0 \\
\addlinespace\multicolumn{3}{l}{\emph{Elementary / response-based components}}\\
Coherent amplitude $a$ & 7/3/1/1/0 & 6/3/1/2/0 \\
Point JAPD & 8/3/1/0/0 & 7/3/1/1/0 \\
Raw 3D median (all pairs) & 8/3/1/0/0 & 5/3/1/3/0 \\
\addlinespace\multicolumn{3}{l}{\emph{Adapted distance and matched support}}\\
M3C2 distance q95 & 6/3/1/2/0 & 5/3/1/3/0 \\
Raw median (M3C2 support) & 8/3/1/0/0 & 6/3/1/2/0 \\
\bottomrule\end{tabular}

\caption{Matched local comparisons: TP/TN/FP/FN/A, twelve requests per column. The strategy, component and adapted-distance groups have different roles. All use the same supplied reconstruction pairs and equal static access. M3C2-support median uses only its usable points; all-pair median does not. Full controls, graph-feature adapters and paired decisions remain in Supplement S6.}\label{tab:published}
\end{table*}

Local-only corrections concern static f01, not missed positives. They reveal different error patterns without establishing a reliable router. Against the M3C2 wrapper, RDSC detects one more positive per regime with the same false alarm. Yet M3C2 retains only 84.93--91.10\% usable points per window; its same-support median improves the inherited-camera count relative to all-pair median. Support selection therefore affects this comparison. Graph-feature extensions establish neither a general accuracy advantage nor evidence against complete published M3C2-LOG.

The broader utility studies test whether added complexity earns its place. Across five error profiles and 26,400 generated pairs, RDSC fails the predefined combination of balanced-loss gain 0.01 and wins in eight of ten banks; in the correlated, heterogeneous profile, the favorable pooled gain is 0.000391. A learned selector does not improve on the best fixed choice, and a separate nonlinear local-coordinate candidate also fails its predefined criteria (Supplement S6--S7).

Fallback has a different, conditional value: coverage. Historical adaptive-window evaluation with inherited cameras gives BASC-C the same four TP, two TN and one FN as RDSC; removing fallback yields seven abstentions. With false-negative cost one, answering has lower loss exactly when abstention cost exceeds $1/7$. This supports fallback under the stated costs, without a general accuracy advantage.

\subsection{RQ4: computational cost and usable inputs}
The cost comparison starts after all required reconstructions exist. Under that condition, RDSC takes about 3.7 CPU seconds versus 13.0 for projected residual scoring (Table~\ref{tab:resources}). BASC-C's all-evidence reference takes 54.2--60.5 seconds and up to 1416 mebibytes (MiB). These are post-SfM prototype costs on the same machine, excluding local reconstruction and optimized lazy-cascade latency.

\begin{table*}[!htbp]\centering\small\setlength{\tabcolsep}{5pt}
\begin{tabular}{llrr}\toprule
Readout & Local cameras & CPU (s) & RSS (MiB)\\\midrule
Projected residual & no locals required & 12.99 & 337\\
RDSC & independent & 3.72 & 273\\
RDSC & inherited & 3.70 & 272\\
BASC-C reference & independent & 60.48 & 1416\\
BASC-C reference & inherited & 54.22 & 1416\\
\bottomrule\end{tabular}

\caption{Post-SfM prototype cost: median process CPU seconds over twelve primary acquisitions per row and maximum process RSS, including imports. Local models must already exist; BASC-C computes every expert. Wall, load/score, machine and co-scheduling details are in Supplement S8.}\label{tab:resources}
\end{table*}

Physical evaluation first requires eligible reconstructions. Known-pose C3VD controls~\cite{c3vd} use privileged geometry; StereoMIS surgical-event labels~\cite{stereomis} supply no displacement truth. In the raw C3VDv2 challenge~\cite{c3vdv2}, available globals register only two to four of ninety images, below the required 72. All methods abstain before local inference (Supplement S7.3). This therefore tests input availability, not detector discrimination or physical-motion accuracy.

\section{Discussion}
\subsection{Interpreting a negative or a positive screen}
The ambiguity in Eq.~\eqref{eq:ray} gives a case in which no statistic of the given tracks separates static and moving geometry. Even when a change leaves a pixel signal after nuisance removal, a test can discard it by restricting the alternative. The localized-family null directions make this second limitation concrete. The conditional detection limit describes one direction under stated geometry and noise assumptions. It does not certify that unseen motion is absent, and a calibrated positive can also arise from camera or reconstruction error. A practical interpretation of a negative decision should identify which motions and amplitudes the acquisition could have revealed.

Keeping the observable point directions makes two previously unscored StereoMIS clips eligible and preserves the original full-rank scores. The focal analysis further separates an invariant calculation at fixed geometry from stability of a fitted reconstruction. Neither changes the amount of independent evidence in a sequence. The original static threshold is particularly sensitive to transfer. On the TUM subset the simpler magnitude scores are better at ranking both scene motion and geometry error. This result and the original-export factorial limit the case for a more elaborate score. The local comparisons similarly retain conditional gains and failures, with component scores and support-matched controls often equalling the complete rules.

\subsection{Evidence from extra views}
An additional view helps in the exact construction because it observes the same state from a distinct centre. The synchronized physical replay tests this idea with sensor depth, and its advantage over maximum separation depends on the candidate cameras and match support. The best fixed view already attains the measured oracle. The endpoint-time offsets and limited registration in the endoscopic tests illustrate why more frames or an optical depth sensor alone are insufficient for a matched tissue-displacement test. This evidence supports a narrower use of the analysis: distinguishing numerical eligibility, physical sensitivity and score transport when deciding how much weight to give a screen.

\section{Conclusion}
Identical monocular tracks can conceal physical change even under exact perspective projection. For a specified geometry and noise model, the conditional gains quantify which motion directions reach the test and how large a displacement must be for a given detection probability. The rank-aware calculation extends numerical coverage, and the added comparisons identify settings where simpler scores match or outperform the modal score. A synchronized view supplies useful metric evidence when the camera and support resolve the ambiguity. Neither the transferred cutoff nor adaptive view selection shows a general advantage in these data. These results give concrete criteria for interpreting a reconstruction screen and for establishing which further observations could resolve an ambiguous negative.

\section*{Statements and Declarations}
\noindent\textbf{Funding.} This research received no specific public, commercial or not-for-profit funding.

\noindent\textbf{Competing interests.} The author declares no competing financial or non-financial interests.

\noindent\textbf{Author contributions.} \'I\~nigo Cirauqui-Viloria conceived the study, developed the methodology and software, performed the investigation, analysis and validation, curated the data, prepared the visualizations and wrote and revised the manuscript.

\noindent\textbf{Ethics and data reuse.} The study reuses existing synthetic, public phantom, surgical and indoor records under the original providers' access and reuse conditions. It involved no new recruitment, intervention or clinical acquisition. No new ethics approval or exemption is asserted.

\noindent\textbf{Data availability.} The original benchmark and selected-model inputs remain accessible at their cited versions~\cite{benchmark,benchmark_support,screeninginputs}. The 24 November campaign protocols, selected results, verification records and Figure~\ref{fig:observability} inputs are available in version \ScreeningMvaVersion\ of the reproduction deposit~\cite{screeningmva}. Online Resource~2 preserves the historical tables and protocols. Online Resource~3 also contains the campaign records. The archived release gives the original providers and their access conditions for third-party data.

\noindent\textbf{Code availability.} The original screening implementation is available at its cited source revision~\cite{screeningcode}. The campaign drivers, archive verifier and Figure~\ref{fig:observability} renderer are available in the versioned reproduction deposit~\cite{screeningmva}.

\section*{Electronic supplementary material}\label{sec:online_resources}
\noindent\textbf{Online Resource 1 (supplement following the article).} Operator details, experimental protocols, complete comparison tables, independent geometry checks and reproduction instructions.

\noindent\textbf{Online Resource 2 (ancillary file \texttt{ESM\_2.zip}).} Historical tables R01--R25, retained protocols and a reading index for the original submitted edition.

\noindent\textbf{Online Resource 3 (ancillary file \texttt{ESM\_3.zip}).} 24 November observability and comparison protocols, source scripts and compact results.

\small
\raggedright
\setlength{\bibsep}{1pt}
\interlinepenalty=10000

\end{document}


\maketitle
\noindent\textbf{Reading route.} S1 defines the global structure-from-motion (SfM) diagnostic and S2 reports its inputs, results and verification. S3--S4 cover local measurements and decisions. S5--S8 document external comparators, local results, utility and coverage limits, and profiling. S9 gives reproduction instructions. S10 indexes Online Resource~2, which contains the 25 earlier tables (R01--R25) and preceding protocols. S11--S14 cover conditional power, attribution and transport, real-video comparisons and synchronized view choices. Online Resource~3 provides their numerical and code records.

\section{Projected residual operator: implementation}\label{sec:soperator}\label{sec:suppglobal}
\subsection{Mode construction and numerical ranks}
Main Figure~1 illustrates how this operator uses all tracks and observations to score one acquisition in the supplied signed-pinhole model. The original COLMAP export consists of \texttt{cameras.txt}, \texttt{images.txt} and \texttt{points3D.txt}. The estimated 90th-percentile (q90) extent normalizes point coordinates and camera translations; intrinsics are fixed. Image names define ordered, evenly spaced times in $[-1,1]$, with linear and squared time. Ten spatial features (constant, three coordinates, three squares and three $\sqrt2$ cross-products), two temporal features and three world axes produce sixty projected columns. Their normalization divides by pixel root mean square (RMS) over observations. The first twenty-four constant/linear spatial columns form affine nuisance; the remaining thirty-six are the alternative. This split can remove genuine affine motion and retain quadratic error patterns. In all scored core models, ten affine directions remain after point and camera elimination, and thirty-six quadratic directions remain after affine removal. The basis is fixed independently of target outcomes, tissue laws and motion trajectories.

Full QR factorization (orthogonal and upper triangular factors) eliminates each point block. Its three singular values must satisfy $s_{\min}^2>10^{-12}s_{\max}^2$. Remove six first-camera coordinates and the largest-magnitude nonzero relative-translation coordinate of the second camera (the scale gauge); coincident/invalid gauge cameras refuse the model. After point elimination, accumulate the camera Gram matrix $S$, the mode-data Gram matrix $T$ and their cross matrix $R$. The sixty mode columns and the residual column give $T$ its $61\times61$ dimensions. Symmetrized camera information must have no eigenvalue below $-10^{-8}\lambda_{\max}$; retained camera directions satisfy $\lambda>10^{-10}\lambda_{\max}$. The camera-null cross-term norm divided by $\sqrt{\operatorname{tr}(S)\operatorname{tr}(T)}$ must not exceed $10^{-8}$. Zero denominators use the smallest positive normal float64 number solely for the diagnostic ratio.

Form $G=T-R^{\mathsf T}S^+R$ and symmetrize. Refuse eigenvalues below $-10^{-7}\|G\|_2$; clip remaining roundoff-negative eigenvalues to zero only for its square root. No positive ridge is introduced. A single physical-information cutoff, $10^{-8}$ times the largest eigenvalue of the sixty-column physical Gram block, determines both affine and remaining quadratic ranks. After affine projection, the retained quadratic left singular vectors define the modal energy $\Delta_{\rm mod}$. With $n_{\rm obs}$ image observations and $n_p$ points,
\begin{equation}
 \nu=2n_{\rm obs}-3n_p-r_C-r_a-q,\qquad
 F_{\rm mod}=\frac{\Delta_{\rm mod}/q}{\mathrm{SSE}_1/\nu}.
\end{equation}
Here $r_C,r_a,q$ are the retained camera, affine and quadratic ranks. In the notation of main Section 3.3, $\mathrm{SSE}_0=\|e_0\|^2$ is the residual sum of squares after nuisance removal and $\mathrm{SSE}_1=\|e_1\|^2$ is what remains after the quadratic fit. Their difference is the modal energy. Zero alternative rank, $\nu\leq0$, or $\sqrt{\mathrm{SSE}_1}\leq10^{-10}\sqrt{r^{\mathsf T}r}$ refuses the studentized readout. Invalid point information refuses the whole model and all matched controls; points are not dropped. The Linear Algebra Package (LAPACK) routine \texttt{gesvd} is a same-equation fallback if \texttt{gesdd} does not converge. Per-model energies, cutoffs, ranks, matrix hashes and refusal reasons are retained.

Pixel median and 95th-percentile (q95) readouts pool the two-dimensional observed-minus-predicted residual norms. Nuisance-null RMS is $\sqrt{\mathrm{SSE}_0/(2n_{\rm obs}-3n_p-r_C-r_a)}$. Each readout has its own maximum-static calibration and strict greater-than decision. Modal energy and the studentized score are two readouts of one operator, not two independently validated methods.

\section{Primary inputs, detailed results and verification}\label{sec:sprimary}
\subsection{Experimental point preparation (EPC)}

EPC means estimator-polished comparison, an experimental input preparation rather than a requirement of the global diagnostic. The fixed selected-input deposit (version \ScreeningInputVersion) supplies the reported models; main Section~6.2 distinguishes their earlier benchmark provenance from current collection access. The original-export interface is evaluated separately in S2.3. The primary input preparation uses the deterministic point solver in \texttt{experiments/paper1\_\allowbreak completion\_\allowbreak 20261023/geometry.py}, SHA-256
\begin{flushleft}\small\ttfamily
70bf384d49b292ffdbfe5c0e333ae1783309bb7f5431368ed73bf8cb0a98c3c7.
\end{flushleft}
For each original point, minimize $c_i(X)=\sum_{o\in\operatorname{unique}(\mathrm{track}_i)}\|\pi(\hat C_o,X;\hat K_o)-y_{io}\|_2^2$ over its three coordinates only, with no robust loss. Tracks, observations, feature identities, camera poses and pinhole intrinsics stay fixed. Global points initialize at the retained global estimate; separate-camera local points initialize at their retained local estimate. Inherited-camera local points initialize after the S3.1 camera-derived similarity and use exactly the corresponding global cameras. No point-truth alignment, pose refinement or track replacement occurs. Thus the global and separate local estimates are also polished here, unlike S6.2's unpolished-global historical comparison.

The damped Gauss--Newton solver uses at most 80 iterations, initial relative damping $10^{-6}$ and steps $-(J_i^TJ_i+\lambda_i h_i I)^{-1}J_i^Tr_i$, where $h_i=\max\operatorname{diag}(J_i^TJ_i)$ (floored at the smallest positive normal). Accepted steps divide damping by three, with floor $10^{-15}$; rejected steps multiply it by ten, capped at $10^{15}$. The stopping criterion is $\|J_i^Tr_i\|_\infty/[\sqrt{h_i}\max(\sqrt{c_i},1)]\leq10^{-7}$. A cost-decreasing step is accepted; a roundoff-neutral step additionally requires strictly decreasing normalized gradient and cost increase at most $32\epsilon p_{\max}\max(\sqrt{2n_ic_i},1)$, with $p_{\max}\geq1$ bounding observed pixel magnitudes. Signed nonzero depth must exceed $10^{-12}$ in absolute value; this fit is not transfer's positive-depth test.

\begin{figure}[tbp]\centering
\includegraphics[width=\linewidth]{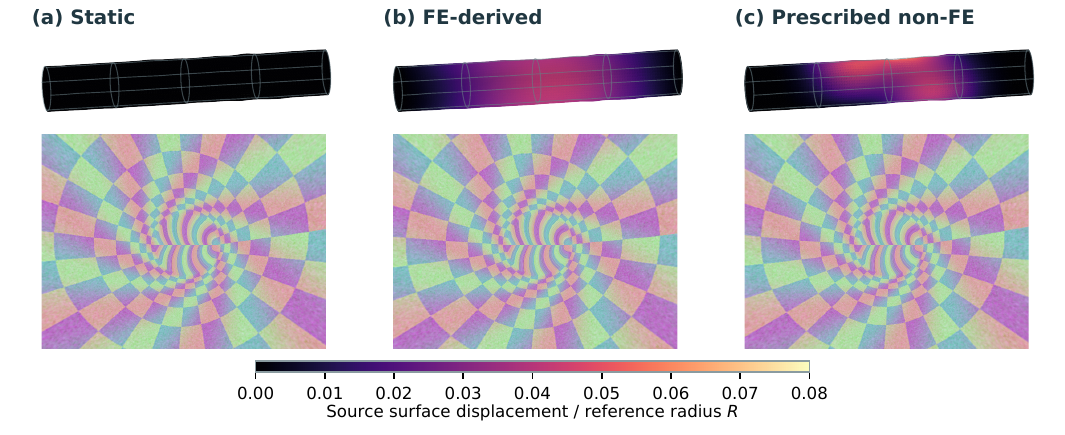}
\caption{Source f01 conditions at frame 52. Top: stored static, finite-element (FE)-derived and prescribed non-FE surfaces at true deformation scale; color is displacement magnitude divided by f01 reference radius $R$, gray wireframe the rest wall. Bottom: retained red, green and blue (RGB) images at the same pose with a common square-root transform. Different prescribed positive fields can produce similar images. These evaluator/source views explain labels, not inference scores; FE-derived kinematics do not establish tissue-mechanical realism.}\label{fig:motionviews}
\end{figure}

The f01 motion conditions in Figure~\ref{fig:motionviews} illustrate the inputs. Fit success requires finite coordinates, at least two usable observations, stationarity and smallest point-information eigenvalue greater than $10^{-12}$ of the largest. The local primary comparison retains the original admission flag and requires at least 100 unambiguous pairs, reference extent above $10^{-12}$, raw q95 discrepancy/$L\leq0.15$, and convergence of every matched global/local point. Failed requested windows remain unavailable; no failed pair is deleted. For the EPC-prepared input condition, every supplied global point fit must converge before global scoring; the diagnostic retains its separate whole-model numerical guards. All 216 EPC window/regime requests are scored. Frozen source/input manifests and the 120 global/separate-local plus 108 inherited-local point arrays are retained; replay consumes these unchanged arrays unless explicitly requesting upstream fit reproduction.

\subsection{Complete core and transfer support}
The primary cohort comprises twelve unique EPC models (729,539 points); transfer comprises twenty unique positive full-cycle exports from f01/f04. The accounting levels are:
\begin{description}
\item[Models:] 12 primary + 20 transfer = 32 unique requested global models.
\item[Calibration definitions:] 12 primary targets $\times$ (full + 3 delete-one calibrations) = 48 target/calibration definitions.
\item[Readout decisions:] 48 definitions $\times$ 5 readouts = 240 primary checks; 20 transfer models $\times$ 5 readouts = 100 transfer checks.
\item[Stored audit records:] the 240 primary answers are repeated under two local-regime labels (480 records). Adding 100 transfer decisions gives 580 records representing 340 distinct checks, not extra acquisitions.
\end{description}
Ninety transfer aliases add no unique models. The cohorts use previously seen anatomy, and transfer contains only positives. Table~\ref{tab:support_summary} summarizes observation support. Tables~\ref{tab:global_all_scores} and~\ref{tab:global_support} list the primary scores and transfer requests.

\begin{table}[htbp]\centering\small\setlength{\tabcolsep}{3pt}
\begin{tabular}{lrrrr}\toprule
Cohort & $n_p$ & $n_{\rm obs}$ & $r_C/r_a/q$ & $\nu$\\\midrule
Core & 57,278--62,654 & 409,369--446,932 & 533/10/36 & 633,044--705,323\\
Transfer & 16,181--50,210 & 134,397--373,084 & 533/10/36 & 217,377--594,959\\
\bottomrule\end{tabular}

\caption{Complete core/transfer support ranges, not independent sample counts. All models have 90 registered images at 640$\times$480 pixels. $n_p$ is requested points and $n_{\rm obs}$ counts two-coordinate observations. Ranks and residual degrees of freedom $\nu=2n_{\rm obs}-3n_p-r_C-r_a-q$ cover scored models only; the transfer rank refusal retains its requested support and has unavailable ranks/df.}\label{tab:support_summary}
\end{table}
\texttt{PRIMARY\_\allowbreak GLOBAL\_\allowbreak SUPPORT\_\allowbreak 20261025.csv} and the accompanying metadata file give every model's dimensions, observation/point support, track-length range/mean, $r_C,r_a,q,\nu$, status, five scores, cutoffs and decisions. Modal energy is summed pixel$^2$; $F$ is dimensionless; median/q95/null RMS are pixels. Mean track lengths are 6.599--7.220 (core), 6.992--9.110 (transfer). Fixed thresholds were tested on these distributions, not controlled support/resolution changes; support-size invariance is unestablished.

\begin{table}[htbp]\centering\small\setlength{\tabcolsep}{4pt}
\begin{tabular}{lrrrrr}\toprule Sequence & $\Delta_{\rm mod}$ & $F_{\rm mod}$ & Median & q95 & Null RMS\\\midrule
f01 / static & 89.526 & 18.925 & 0.22028 & 1.0177 & 0.36267 \\
f01 / FE-derived & 715.79 & 148.05 & 0.22656 & 1.0301 & 0.36784 \\
f01 / prescribed & 3472.4 & 590.24 & 0.26666 & 1.1612 & 0.4104 \\
f02 / static & 82.522 & 16.547 & 0.2306 & 1.0462 & 0.37236 \\
f02 / FE-derived & 620.16 & 122.42 & 0.2364 & 1.0557 & 0.37638 \\
f02 / prescribed & 4713.5 & 787.03 & 0.27291 & 1.1767 & 0.41671 \\
f03 / static & 77.942 & 15.608 & 0.23785 & 1.0258 & 0.37259 \\
f03 / FE-derived & 2948.8 & 542.63 & 0.25964 & 1.0944 & 0.39426 \\
f03 / prescribed & 7359.8 & 1041.2 & 0.30575 & 1.2631 & 0.45604 \\
f04 / static & 57.868 & 11.902 & 0.22703 & 1.0232 & 0.36762 \\
f04 / FE-derived & 308.69 & 62.393 & 0.23159 & 1.0302 & 0.37134 \\
f04 / prescribed & 5406.1 & 925.46 & 0.274 & 1.1455 & 0.41302 \\
\bottomrule\end{tabular}

\caption{Complete twelve-model global scores, rounded for display only. Modal energy is summed squared projected pixel residual; median/q95/null RMS are in pixel units and $F_{\rm mod}$ is dimensionless. Cutoffs, decisions and every calibration-deletion result use stored full-precision values. These are EPC-refined core coordinates, not retrospectively relabeled original-export scores.}\label{tab:global_all_scores}\end{table}

In the primary and original transfer tests, each target uses the maximum of the three static scores from other families. The primary test also retains every specified delete-one fold. A unique static maximum forces one false alarm, not population specificity or a nominal $F$ result. Original-export calibration is recomputed from original scores rather than mixed with EPC cutoffs.

\begin{table}[htbp]\centering\small\setlength{\tabcolsep}{4pt}
\begin{tabular}{llllrr}\toprule Family & Appearance & Pipeline & Status & Points & $q$\\\midrule
f01 & diffuse & A & scored & 30,947 & 36 \\
f01 & diffuse & B & scored & 29,350 & 36 \\
f01 & coated & A & scored & 18,144 & 36 \\
f01 & coated & B & scored & 17,412 & 36 \\
f01 & asymmetric & A & scored & 30,455 & 36 \\
f01 & asymmetric & B & scored & 28,385 & 36 \\
f01 & half exposure & A & scored & 17,321 & 36 \\
f01 & half exposure & B & scored & 16,632 & 36 \\
f01 & double exposure & A & scored & 50,210 & 36 \\
f01 & double exposure & B & scored & 47,440 & 36 \\
f04 & diffuse & A & scored & 29,439 & 36 \\
f04 & diffuse & B & scored & 27,708 & 36 \\
f04 & coated & A & scored & 17,704 & 36 \\
f04 & coated & B & scored & 16,946 & 36 \\
f04 & asymmetric & A & rank abstention & 29,434 & -- \\
f04 & asymmetric & B & scored & 27,221 & 36 \\
f04 & half exposure & A & scored & 16,881 & 36 \\
f04 & half exposure & B & scored & 16,181 & 36 \\
f04 & double exposure & A & scored & 48,466 & 36 \\
f04 & double exposure & B & scored & 45,355 & 36 \\
\bottomrule\end{tabular}

\caption{Every unique appearance/pipeline request. All register ninety images. The rank-refused model retains its 29,434 requested points and is an abstention for all five matched global readouts. No replacement reconstruction, track deletion or numerical-floor adjustment is used.}\label{tab:global_support}\end{table}

Transfer has 19 true positives (TP), zero false negatives (FN) and one abstention (A) for each modal readout; median/q95/null RMS give 3/16/1, 8/11/1 and 5/14/1. All five readouts count all twenty requests, including the abstention. Appearance specificity and physical error remain unevaluated.

\subsection{Original exports and implementation verification}
We ran two replays on the same 32 models. Using the primary inputs reproduced all 32 score dictionaries bitwise from 108 files (36 core global text files, twelve core point arrays and sixty transfer text files). Using only the ninety-six original text files and recalibrating on original static scores reproduced all 340 decisions (240 core and 100 transfer), together with their ranks, statuses and failure reasons. The 580 stored records include the duplicate core entries described in S2.2. Transfer scores remain bitwise equal; core scores change by up to 0.479259796316 across readout scales. These replays test decision agreement on the supplied models. Changes in support, resolution or reconstruction and reproduction from the original images are outside their scope.

Four 4,096-point retained qualification controls compare streamed/direct Schur Grams and independently solved least squares: maximum Gram-entry error $4.720\times10^{-10}$, affine/quadratic ranks 10/36. Core audits check twelve matrices and 1,248 answers on Ubuntu/macOS, with maximum statistic discrepancies $5.821\times10^{-11}$/$7.320\times10^{-9}$. Transfer audits check nineteen matrices, one hundred decisions, sixty input identities and ninety aliases. Twelve core/six transfer macOS library warnings remain logged; finite results pass fixed tolerances.

S9 gives runtime, scalar/Gram tolerances and backend limitations. The scientific capsule and R21--R23 protocol context retain complete matrices, hashes and logs. These checks qualify execution, not physical validity.

\subsection{Optional normalization and predefined transfer criteria}
The $F_{\rm mod}$ readout uses the same projected operator as modal energy. Its predefined improvement criterion required an extra correct answer over every global control, including energy. It fails because the two readouts produce identical decisions. A separate criterion for the common operator required at least eighteen scored transfer requests and sixteen detections per modal readout; both reach nineteen. These outcomes concern different questions and remain separate from the earlier nonlinear local-coordinate candidate in S7.2.

\section{Global--local expert definitions}\label{sec:sexperts}
Table~\ref{tab:experts} summarizes the inputs and measurements of the local experts. Each measures reconstruction consistency.
\begin{table}[tbp]\centering\small
\begin{tabularx}{\linewidth}{lXX}\toprule
Expert & Measurement & Additional evidence\\\midrule
RDSC & Robust detrended spatial coherence of point differences & Matched tracks, aligned points, neighborhoods\\
Transfer & Global/local middle-view prediction contrast & Shared observations and endpoint triangulation\\
Raw q95 & Magnitude of geometric disagreement & Aligned point differences\\
JAPD & Joint adjustment-propagated disagreement of point positions & Shared point/camera perturbation responses\\
Squared-edge & Neighboring squared-length differences & Neighborhoods and joint responses\\\bottomrule
\end{tabularx}
\caption{Global--local expert orientation. Exact scalars, support conditions and missing values are defined below.}\label{tab:experts}
\end{table}

\subsection{Coordinates, identities and support}
A track is the set of image/keypoint identities attached to an estimated 3D point. A global/local point pair is a candidate if its tracks share at least three distinct observation identities. Retain it only when neither point has another candidate match; do not resolve ambiguity using 3D proximity or motion truth. The two models must share image identities and the same feature indexing within a frontend.

\begin{figure}[tbp]\centering
\includegraphics[width=\linewidth]{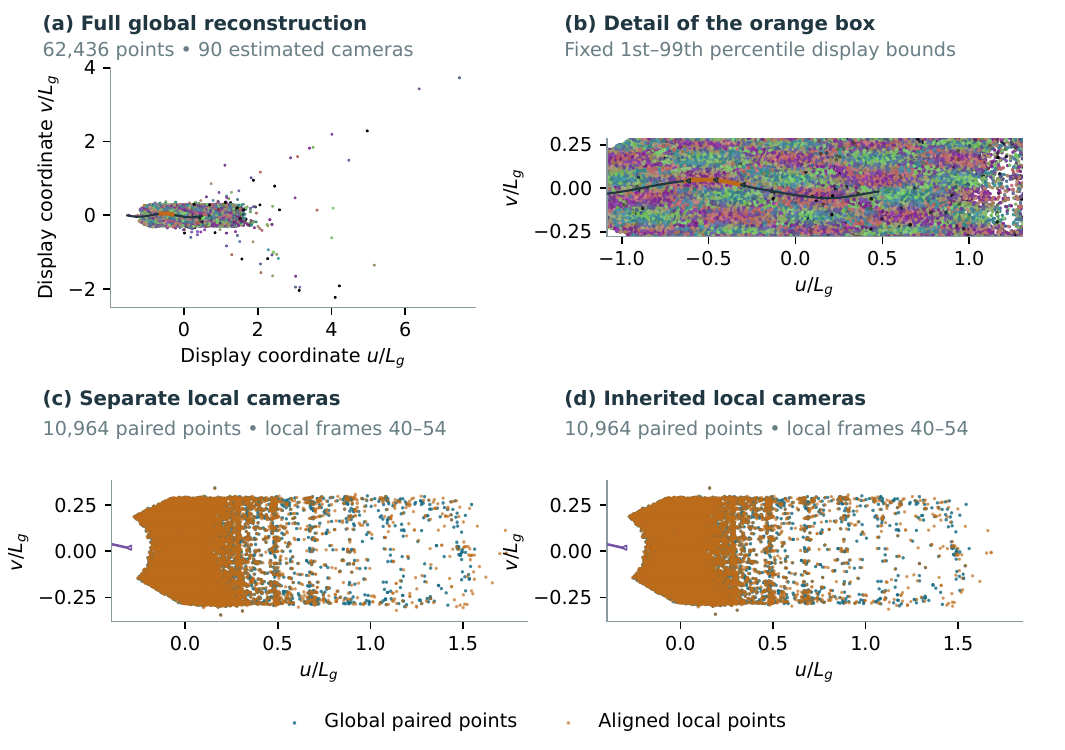}
\caption{Global/local SfM evidence, f01/finite-element (FE)-derived motion: deformation bases come from finite-element solves; motion timing is prescribed, not physiological dynamics. (a) All 62,436 global points/outliers and 90 estimated camera centers; dark global path, orange selected window. (b) Fixed 1st--99th percentile detail bounds, marked in (a). (c,d) All 10,964 matched pairs, window 4/images 40--54, with separate/inherited cameras and unchanged camera-derived similarity alignment; purple shows aligned local paths/orientations. Paired panels share bounds, without surface/truth alignment. Coordinates share a rotation and estimated extent $L_g$; fixed-depth glyphs are not calibrated fields of view. Point colors: retained export in (a,b), teal global/orange local in (c,d).}\label{fig:bundles}
\end{figure}

Write camera orientations as world-to-camera rotations $R^g_j,R^l_j$ and centers as $c^g_j,c^l_j$. For at least two shared cameras with matching image names, compute the singular-value decomposition $U\Sigma V^{\mathsf T}=\sum_j(R^g_j)^{\mathsf T}R^l_j$ and set
\begin{align}
 R_a&=U\,\mathrm{diag}(1,1,\det(UV^{\mathsf T}))V^{\mathsf T},\\
 s&=\frac{\sum_j(c^g_j-\bar c^g)^{\mathsf T}R_a(c^l_j-\bar c^l)}{\sum_j\|c^l_j-\bar c^l\|^2},\qquad
 t_a=\bar c^g-sR_a\bar c^l.
\end{align}
The bars denote camera-center means. Reject a denominator at most $10^{-18}$ or a nonfinite/nonpositive scale. The alignment $T(y)=sR_a y+t_a$ is estimated from corresponding camera poses and is never optimized to reduce the point discrepancies subsequently evaluated. Figure~\ref{fig:bundles} shows the aligned models under both camera regimes. For matched global points, use $L=q_{.90}(\|x_i-\bar x\|)$, with $\bar x$ their centroid. All quantiles use linear interpolation between ordered values.

Ordinary point statistics require 100 pairs, $L>10^{-12}$ and raw q95/$L\leq0.15$. Later camera-release/physical collections retain their own frozen support/admissions; comparisons are matched within each collection. The decision cascade, Bundles applicability-aware selective composition with cascade preservation (BASC-C), preserves the original inputs and requested windows. Missing models, failed admission and numerical failures remain unavailable scores.

\subsection{Original RDSC}
RDSC removes an affine trend, measures coherent neighborhood residuals and compares them with shuffled arrangements before combining the two amplitudes.

Let $d_i=T(y_i)-x_i$ and let row $D_i=[(x_i-\bar x)^{\mathsf T}/L,1]$ form an $n\times4$ affine design, and let $B$ be its $4\times3$ coefficient matrix. Starting from unit weights, fit the three displacement coordinates jointly by weighted minimum-norm least squares. At each of at most 20 iterations, set
\begin{align}
 r_i&=\|d_i-D_iB\|,\\
 \sigma&=\max\{1.4826\operatorname{median}_i|r_i-\operatorname{median}_j r_j|,\epsilon L\},\\
 w_i&=\min\{1,1.345\sigma/\max(r_i,\eta)\}.
\end{align}
The stopping tolerance on weight change is $10^{-8}$. Here $\epsilon$ is float64 machine precision and $\eta$ its smallest positive normal number. For the fitted residual vectors, use the nine entries returned by a nearest-neighbor query at each reference point (normally the point itself and eight neighbors) and compute the q95 norm of neighborhood means divided by $L$. The shuffled reference repeats the same pooling after 20 residual-identity permutations and takes their median. The NumPy default generator uses seed $20260904+t$ for retained window index $t$. Original point-row order and tied-neighbor behavior are preserved; this component does not claim arbitrary row-order invariance. The final score is coherent amplitude squared divided by the floored shuffled amplitude. Raw median and raw q95 use the original displacement norms divided by $L$.

\begin{figure}[!htbp]\centering
\includegraphics[width=\linewidth]{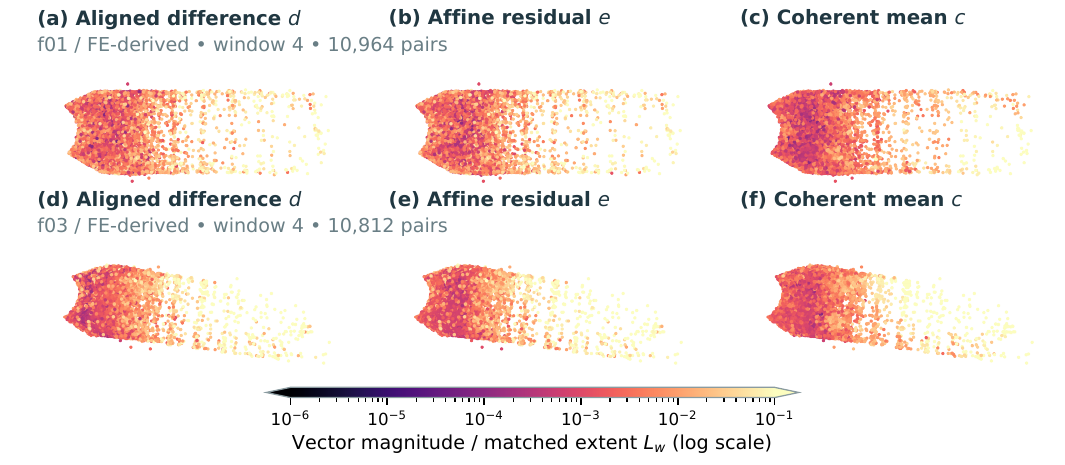}
\caption{RDSC intermediates: window 4, f01/f03 FE-derived acquisitions, separately estimated local cameras, all matched points. Color shows norms of aligned discrepancy $d$ (a,d), robust affine residual $e$ (b,e), or nine-entry neighborhood mean $c$ (c,f), divided by original-export matched extent $L_w$. Stages share coordinates/bounds within each separately centered row. The common logarithmic scale saturates outside $[10^{-6},10^{-1}]$ for display only; vector directions are not shown and scores are unchanged. These are reconstruction discrepancies, not measured material displacement.}\label{fig:rdscfields}
\end{figure}

This affine nuisance removal may also remove real affine deformation. The shuffled score is a finite empirical reference, not a hypothesis-test probability. Figure~\ref{fig:rdscfields} shows the discrepancy magnitude at each step.

\subsection{Three-view transfer intervals}
For each paired point, choose first, middle and last distinct shared observations ordered by image name, with image ID breaking ties. For $v$ shared views, use zero-based indices $0,\lfloor v/2\rfloor,v-1$ (the later middle view when $v$ is even). Shared pixel coordinates must agree between models within $10^{-8}$ pixels; duplicate shared observations within a view are averaged. The retained transfer operator accepts pinhole intrinsics $(f_x,f_y,c_x,c_y)$; unsupported camera models make it unavailable. Under each camera set, unproject endpoint rays and compute the midpoint of their closest points, then project it into the middle view. Normalize camera-center coordinates for the solve. The singular-value degeneracy test uses $64\epsilon$ relative tolerance; valid finite projections and positive depth in all three views are required. The contrast uses squared middle-view pixel errors $g,\ell$: $g/(g+\ell)$, or $1/2$ when both are zero. If either error is unavailable, the point contributes $[0,1]$. Median lower and upper point bounds form the window interval.

The middle observation is excluded from this triangulation operation, but the supplied SfM camera estimates may already have used it. This is a conditional transfer contrast, not an end-to-end held-out image prediction. For exactly identical global/local camera inputs and observations, the two errors are identical; the valid contrast is necessarily $1/2$. Small numerical export differences are separately recorded in the upstream evidence rather than assumed to be new estimator independence.

\subsection{Joint point response}
Joint adjustment-propagated disagreement (JAPD) linearizes 2D reprojection with respect to three point coordinates and six camera-pose coordinates, holding intrinsics fixed. Each model is normalized by its estimated q90 extent. Point elimination uses complete QR rather than subtracting large normal-matrix blocks. One camera pose and a nonzero translation component of a second camera fix the seven similarity gauge directions. Point and reduced-camera information floors are $10^{-12}$ and $10^{-10}$ times their respective largest eigenvalues. These regularizers are numerical devices, not measured priors.

To estimate how shared image errors affect the aligned point difference, we propagate 128 unit random-sign (Rademacher, $-1/+1$) pixel-coordinate perturbations through the two linearized estimators. The covariance below scales these responses by the empirical noise level. Each probe is keyed by image identity, keypoint index, pixel-coordinate component (x/y), probe index and seed 20260924. The key uses the coordinate component, not the numerical observed pixel value. Shared observations therefore receive the same perturbation in both models. All right-hand sides are solved together; reusable global responses are cached. The component uses the analytic differential of the camera-derived similarity and the global-frame reanchoring term. Joint adjustment and gauge treatment follow the geometric foundations cited in the main article; the response construction here couples the two models through shared probes. For inherited local cameras, the local point response is conditioned on the shared global-camera response rather than introducing independently perturbed local camera poses. This includes conditional point, camera and alignment response rather than assuming independent errors for two reconstructions of shared observations.

For response vectors $z_{ij}$, the covariance proxy is $C_i=\sigma^2\sum_j z_{ij}z_{ij}^{\mathsf T}/128$. The empirical scale is
\begin{equation}
 \sigma^2=\operatorname{median}_i\frac{\rho_i}{2-3/n_{\mathrm{local},i}},
\end{equation}
where, for the shared observation-identity set $S_i$ and local residual vector $r^l_{io}=y_{io}-\pi(\hat C^l_o,\hat X^l_i)$, the implemented scale uses
\begin{equation}
 \rho_i=\frac{1}{|S_i|}\sum_{o\in S_i}\|r^l_{io}\|_2^2,
 \qquad n_{\mathrm{local},i}=|\operatorname{unique}(\mathrm{track}^l_i)|.
\end{equation}
Thus $\rho_i$ averages squared two-coordinate vector norms, not individual coordinate components. The denominator $2-3/n_{\mathrm{local},i}$ deliberately retains the full track length used for the three-parameter local point fit, while the residual numerator uses only shared observations. It is an empirical conditional proxy, not an exact shared-subset degrees-of-freedom correction or a demonstrated unbiased noise estimator. With the regularized/floored covariance $\widetilde C_i$ defined below, the actual window scalar is
\begin{equation}
 s_{\mathrm{JAPD}}=\operatorname{median}_i
 \sqrt{\max\{d_i^{\mathsf T}\widetilde C_i^{-1}d_i,0\}}.
\end{equation}
The square root is taken pointwise before the median; there is no division by three. Before inversion, add $\epsilon\max(q_{.90}(\|d_i\|)^2,\eta)I$ and floor eigenvalues at $10^{-12}$ of the largest eigenvalue. An identically zero discrepancy returns zero. Fixed intrinsics, conditional track support, omitted matching/selection uncertainty and empirical noise scaling prevent interpreting the result as a calibrated physical-deformation probability.

\subsection{Second-order squared-edge contrast}
For eight nearest reference-neighbor edges, normalize coordinates and responses by the same fixed reference extent. Write $e_x=x_j-x_i$, $e_y=T(y_j)-T(y_i)$, $m=(e_y+e_x)/2$, and $v=e_y-e_x$. The squared-length difference is $a_0=2m^{\mathsf T}v$. Let $h_{e,r}$ stack the three midpoint and three difference responses of edge $e$ to probe $r$, after extent normalization. Form $B_{ef}=\sigma^2\sum_{r=1}^{128}h_{e,r}h_{f,r}^{\mathsf T}/128$. From this joint covariance of the stacked edge midpoint/difference responses, form the bias-corrected contrast $a_e=a_{0,e}-2\operatorname{tr}(B_{mv,ee})$ and set $g_e=(2v_e,2m_e)$. The edge covariance is
\begin{equation}
 C_{ef}=g_e^{\mathsf T}B_{ef}g_f+
 4\sum_{ij}\left[B_{mm,ef,ij}B_{vv,ef,ij}+B_{mv,ef,ij}B_{vm,ef,ij}\right].
\end{equation}
Here $e,f$ index the eight edges, $i,j$ index Cartesian coordinates, and $B_{mm},B_{mv},B_{vm},B_{vv}$ are the $3\times3$ cross-covariance blocks of midpoint and difference responses. The second term retains uncertainty when an edge has zero length. The point score is $\sqrt{a^{\mathsf T}C^{-1}a/8}$ and the window score is its q95. Add $[(64\epsilon)^2+\epsilon\operatorname{mean}(a^2)]I$ and floor eigenvalues at $10^{-12}$ of the largest eigenvalue. Gaussian moment identities justify this conditional algebra; the observed SfM errors have not been shown Gaussian or unbiased.

Graph ties within relative tolerance $10^{-12}$ use stable lexicographic ranks of sorted unique image/keypoint track signatures. Duplicate signatures are unavailable. Coincident points remain included. This identity-based tie rule defines the squared-edge comparison graph. Neither graph construction nor edge differences estimate material strain. Rigid local motion and isometric changes can be suppressed.

With inherited cameras, both geometric components must supply a decision. If either is unavailable, the geometric expert is unavailable. Otherwise it returns positive if either component is positive and negative if both are negative.

Both components use the same adjustment responses. JAPD scores point positions, while squared-edge scores relative geometry between neighbors.

\section{Window evidence, decisions and composition}\label{sec:spooling}
\subsection{Window evidence and missingness}
A window record identifies a feature-extraction/matching pipeline (frontend arm), requested temporal index and either an ordered finite interval or a missing value. Repeated arm/index identities are rejected. At each index, the maximum available lower bound and maximum available upper bound are used. Unavailable indices are retained in coverage accounting.
\subsection{Temporal pooling}
At least $\lceil0.8N\rceil$ of the $N$ requested indices and a contiguous run of length $k$ are needed. Transfer uses $k=3$; every other supplied expert uses $k=2$, including RDSC and scalar controls in historical appearance contracts. BASC-C routes these sequence-level decisions; it has no additional temporal persistence. The primary EPC schedule and historical schedules are distinguished in Table~\ref{tab:temporal_policy}. For each bound, take the maximum over run minima. Missing indices do not become adjacent.

\begin{table}[htbp]\centering\small\setlength{\tabcolsep}{3pt}
\begin{tabularx}{\linewidth}{>{\raggedright\arraybackslash}p{.34\linewidth}Xrr}\toprule
Cohort / expert & Requested schedule and coverage denominator $N$ & $k$ & Minimum\\\midrule
EPC: transfer & Nine 15-image windows; starts $0,10,\ldots,70,75$; $N=9$ & 3 & 8/9\\
EPC: RDSC, JAPD, edge, raw and scalar controls & Same nine windows, both camera regimes & 2 & 8/9\\
Historical tubular / appearance: transfer & Nine retained indices $0,\ldots,8$; $N=9$ & 3 & 8/9\\
Historical tubular / appearance: other experts & Same requested anchors, including unavailable ones & 2 & 8/9\\
Historical C3VD / curved: transfer & C3VD $N=4,9,10$; curved $N=18$ & 3 & $\lceil.8N\rceil$\\
Historical C3VD / curved: other experts & Per-sequence indices $0,\ldots,N-1$ & 2 & $\lceil.8N\rceil$\\
Physical collections: transfer / other experts & Nine selected-image anchors; pool arms per index, $N=9$ & 3 / 2 & 8/9\\
Global modal: core / transfer & One full 90-view model; no local-window pooling & -- & model gate\\\bottomrule
\end{tabularx}
\caption{Executed expert-by-cohort persistence and requested denominators. All local endpoints maximize run minima on consecutive ordinal indices; missing indices never become adjacent. Coverage counts unique requested anchors after pooling frontend arms, not available windows or arm records. Physical cohorts include the Colonoscopy 3D Video Dataset (C3VD). Fixed physical windows start at $0,9,\ldots,72$ and adaptive lengths follow S6.1; EPC instead uses the retained 10-image stride plus final window. \texttt{TEMPORAL\_\allowbreak CONTRACTS.json} gives every group/expert index and denominator, and all primary window image indices were verified against the exports. Historical per-window retained image identities and registration caveats are in \texttt{LEGACY\_\allowbreak IMAGE\_\allowbreak IDENTITIES.json}.}\label{tab:temporal_policy}
\end{table}
\subsection{Static calibration and abstention}
For each target and expert, the calibration interval is the componentwise maximum of all designated static sequence intervals. If the list is empty or any required endpoint is missing, calibration is unavailable. A target is positive only if its lower bound exceeds the calibration upper bound, negative if its upper bound does not exceed the calibration lower bound, and otherwise abstains. For scalar scores, equality gives a negative screen. For intervals, the endpoint comparisons above determine whether the evidence is decisive. These intervals are conditional on admitted windows and available frontends. They are not bounds on the hypothetical endpoint of a completed schedule and are not confidence intervals.
\paragraph{Why a missing window matters.}
For example, with nine requested anchors and persistence three, known scores $1,1,\bot,0,0,0,0,0,0$ satisfy 80\% coverage. Available complete runs give endpoint $[0,0]$, but filling the missing anchor with one would give $[1,1]$. Pointwise transfer indeterminacy and missing whole windows are different forms of missingness.

\subsection{Provenance-aware composition}
\begin{figure}[!htbp]\centering
\includegraphics[width=\linewidth]{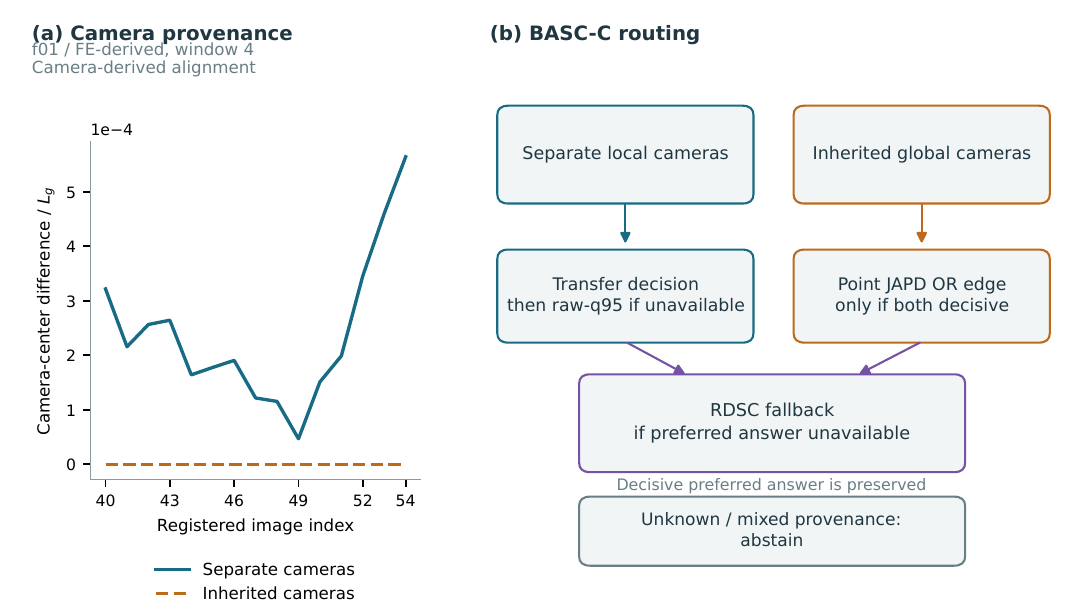}
\caption{Camera provenance and BASC-C. (a) f01/FE-derived window 4 local camera-center differences after camera-derived alignment and $L_g$ normalization. Shared inputs make the inherited-camera curve zero to roundoff; the separate-camera curve shows reconstruction difference, not pose error. (b) Fixed branches route calibrated sequence decisions, preserving decisive preferred answers and using RDSC only when unavailable. Unknown/mixed provenance abstains; global modal is separate from the cascade.}\label{fig:bascdiagram}
\end{figure}

For BASC-C, let $d_T,d_Q,d_J,d_E,d_R\in\{0,1,\bot\}$ be calibrated transfer, raw-q95, JAPD, squared-edge and RDSC decisions. Unavailable decisions are $\bot$. The preferred answer is selected by applicability, not estimated accuracy:
\begin{equation}
 d_P=\begin{cases}
 d_T,&\text{separate cameras and }d_T\ne\bot,\\
 d_Q,&\text{separate cameras and }d_T=\bot,\\
 d_J\lor d_E,&\text{inherited cameras, both decisive},\\
 \bot,&\text{inherited cameras, either unavailable}.
 \end{cases}
\end{equation}
For known homogeneous camera provenance, the cascade returns $d_P$ if decisive and otherwise $d_R$. Unknown/mixed provenance abstains. A decisive preferred answer, including a wrong one, is preserved. The global diagnostic is evaluated separately. No-R omits the RDSC fallback and No-Q omits the raw-q95 step. With known homogeneous camera provenance, the full cascade can answer wherever No-R or RDSC can answer. Figure~\ref{fig:bascdiagram} shows these routes.

Qualification covers 1,539 routing/preservation/coverage assertions and malformed inputs. Exact pseudocode remains in the supporting protocol record.

\section{External comparator adaptations}\label{sec:scomparators}
\subsection{Matched native M3C2 distance adaptation}
Multiscale model-to-model cloud comparison (M3C2) and the other comparators use the same 216 exposed EPC window/regime point-pair records and all 96 target/calibration definitions. Operators, parameters, support and pooling are frozen before feature scoring. No new reconstruction, optimization or input cohort is introduced. Table R25 preserves all deletion outcomes.

The M3C2 operator uses the official native py4dgeo 1.2.0 ARM64 wheel, SHA256 \texttt{e248263aca078cc8\allowbreak db1fca3c1fac2ed2\allowbreak dc36c4a1f86d5480\allowbreak 4eb65e409168c8ba}. An isolated CPython 3.12 environment pins NumPy 2.2.6/SciPy 1.16.1 and retains the full installation resolution and native source/binary hashes. The normal frontend is the qualified per-radius covariance calculation from the earlier M3C2 study, rather than that study's failed default KDTree implementation that accumulated neighborhoods between radii. The failed Octree/default attempts remain historical evidence; they are not new benchmark outcomes.

For each reference point and radius 0.05/0.10/0.20 after centroid/q90 normalization, compute covariance about the neighborhood mean with divisor $n-1$. Choose the smallest-eigenvalue normal from the scale maximizing $(\lambda_1-\lambda_0)/\lambda_2$; orient toward positive Z. Invalid planarity or insufficient support returns an invalid normal. Native mean cylinder distances use radius 0.05, half-depth 0.25 and registration-error input zero. Unit normals, finite distances and at least five samples in each epoch define usable points. Require at least 100 pairs, raw normalized q95 at most 0.15, at least 100 usable points and usable fraction at least 0.8. Read out q95 absolute distance, plus same-usable-support raw median/q95 controls. These are empirical sequence features; neither native level-of-detection output nor a zero-uncertainty significance test is used.

Identity, normal/tangential translations, common rigid transform and insufficient-overlap fixtures qualify M3C2. All 216 windows score with usable fractions 0.8492956--0.9110008; independent normals, cylinder counts, distances and decisions agree within fixed tolerances (full audit in the supporting protocol).

\subsection{Adapted LOG-lite features and zero-distance extension}
The author source is pinned at revision \texttt{e7824b14f1823dddead52480f062a9b8cd9f4d1b}; six MATLAB files and README are hash-bound. LOG-lite adapts the location and orientation united graph comparison features. The snapshot has no license file. Its provenance is retained privately, without relicensing or redistribution. The adaptation excludes the full covariance/significance method.

Normalize both paired clouds by subtracting the first estimated global camera center and scaling the original matched extent to 20. The source seed matrix is
\begin{equation}
 M_i=\frac1{K_i}\sum_{j\in\mathcal B_i}\|x_j-x_i\|^2(x_j-x_i)(x_j-x_i)^{\mathsf T}.
\end{equation}
It is a radial-weighted second moment about the query, not ordinary principal component analysis about a neighborhood mean. Start with radius 2 and enlarge by integer factors until at least eleven neighbors are present. Use the smallest-eigenvalue normal, orient toward the origin, and mark a relative eigen-gap at most $10^{-10}$ unavailable. The graph uses self plus three nearest neighbors and inverse-distance weights on the other entries. With normalized neighbor matrix $W$, the two-level feature is $(\frac12I+\frac14W+\frac14W^2)X$, and the same construction applies to normals. Location is the normalized paired feature distance; orientation uses the equivalent stable atan2(cross,dot) angle. The supplied track pairing replaces the source's external correspondence list. Per-window q95 and two-window persistence form separately calibrated sequence readouts. Invalid normal angles remain intervals $[0,\pi]$, with at least 80\% point coverage; whole unavailable windows remain in the 80\% sequence coverage denominator.

The original feature adaptation cannot score any window because inverse-distance weights are undefined for coincident neighbors. Its 216 reasons and all twelve sequence abstentions per regime are preserved. A distinct, subsequently frozen zero-distance (LOG-lite/ZD) adapter assigns uniform weight among zero-distance neighbors and zero elsewhere. No point is deleted, merged or perturbed, and no graph radius or pooling parameter is changed. For exactly identical coordinates/neighborhood normals this resolves feature evaluation without arbitrary finite jitter; it does not assert a unique limit for all unequal vanishing-distance paths. Location and orientation are calibrated independently. The union is their OR only if both are decisive; otherwise it abstains. This is not the full paper's covariance union.

Dense-matrix oracles qualify forty distinct-cloud and 48 ZD fixtures, including duplicate/degenerate cases. Independent post-closure point-feature and decision checks pass; maximum location/angle discrepancies are $8.527\times10^{-14}$/$8.161\times10^{-15}$. Primary/ZD pair hashes agree. Table R25 records full and deletion denominators (12/9), which remain dependent.

\subsection{Scope of unexecuted systems}
The sparse exports lack the sensor/scan positions and propagated registration uncertainty required by M3C2 with error propagation (M3C2-EP). Substituting zeros would change its uncertainty model. We also did not run Flood's full map-merging system or the stability-driven region growth framework for 3D change detection (SRGCD)~\cite{srgcd}. On 24 October the SRGCD authors still listed code and models as forthcoming~\cite{srgcd_code}. These systems have no measured outcomes here.

\section{Complementary global--local results}\label{sec:slocalresults}
\subsection{Historical cohorts and experimental dependence}
The historical ledger has 118 reconstruction requests, 306 calibration definitions, 93 labeled tasks and 2,160 windows (R02--R04), not independent acquisitions. Counts (requests/definitions/labeled/windows) are original 41/41/36/594, fixed/adaptive inherited each 12/12/7/324, released/fixed-input 41/193/36/594, regenerated 12/48/7/324. Each physical collection requests three C3VD calibrators, seven labeled targets and two unknown surgical-event labels. Unknowns remain predictions but are excluded from confusion matrices. Physical class labels are domain-confounded, not point-motion truth.

The four tubular families couple geometry, texture, camera path and motion; f03 doubles the motion bases. Ten appearance conditions reuse FE-derived f01/f04 under diffuse symmetric/asymmetric, coated symmetric and half/double exposure. The two older curved strengths share a different scene/renderer. Legacy C3VD uses estimated cameras under a pinhole approximation, not privileged known poses. R02 preserves every clip/role and R04 every reconstruction/admission profile. The generator and benchmark remain separately cited~\cite{benchmark,lumenmorph}; physical sources are C3VD~\cite{c3vd} and StereoMIS-tracking~\cite{stereomis}.

The physical frontends use the scale-invariant feature transform (SIFT, 12,000 features) or a lighter keypoint and descriptor extraction network (ALIKED, 2,048 features) with LightGlue or mutual-ratio matching; arms pool by maximum per anchor. Nine anchors start at selected-image indices $0,9,\ldots,72$. Fixed locals contain fifteen images. Adaptive radii 7/15/22 around anchor+7 stop at the first support with 100 global points seen in three views and ray separation at least $1.5^\circ$; otherwise the last support remains subject to admission. Releasing cameras refines poses/points under fixed intrinsics; fixed-input replay preserves databases, whereas regeneration changes models. These are input challenges, not score-selected alternatives.

The twelve dependence blocks are f01--f04, the older curved scene, cecum/descending/sigmoid/transverse phantom segments, and surgical subjects P1--P3. Within-block variants stay together; blocks need not be statistically independent. Tubular/appearance calibration excludes the target family; older curved and legacy C3VD use all four synthetic statics. Physical inference uses three C3VD calibrators. Full/delete-one definitions retain missingness; no deletion is selected as a replacement cutoff. R05--R09 preserve strata, removals, deletions and branch attribution. The global modal diagnostic cannot be evaluated on historical point-only inputs lacking pixels; they are not pooled with EPC.

In the original 29-target legacy comparison BASC-C has 26 correct decisions versus nineteen for RDSC; No-R already supplies this advantage. In original physical data No-Q loses one BASC-C positive. Fixed inherited inputs give seven abstentions. In adaptive inherited inputs RDSC fallback adds six correct and one incorrect answer over No-R. Regenerated BASC-C gives five TP and two false positives (FP). RDSC gives four TP, two true negatives (TN) and one FN, hence more correct answers. Dual geometry's ten historical appearance positives have no matched inherited-camera static branch, so branch specificity is untested. Reuse of domains and subjects does not resolve these limits.

\subsection{Sensitivity to camera and calibration conditions}
The earlier conditioning study uses unpolished globals and inherited local point refits, unlike EPC (S2.1). It retains all 216 windows, original tracks and one-to-one three-observation pairs. Its 80-iteration point solver uses the same $10^{-7}$ normalized-gradient criterion and numerical stationarity repair documented in S2.1; the initial 96 refused windows remain recorded. Final components are all available, maximum gradient $9.99994\times10^{-8}$. Conditioning changes point estimates as well as cameras, so it is not a camera-only causal intervention.

Full calibration has twelve targets; deleting a calibrator gives nine targets that actually used it (no no-op replicas). R10 preserves all counts. Inherited geometric components detect only f04 FE/non-FE and falsely flag f04 static; their decisive negatives block otherwise positive RDSC answers. Removing f04 calibration adds five detections and two false alarms without justifying a replacement threshold.

R12--R13 retain the full original-RDSC attribution matrix and exact matched control definitions (similarity/rigid/local-affine fits, edge change, vector Moran, Chamfer and Hausdorff wrappers). The older pose-conditioned replay gives RDSC 6 TP/6 TN versus raw median 2 TP/5 TN/1 FP/4 FN; coherent amplitude already gives 6 TP/5 TN/1 FP. Deleted raw imagery prevents end-to-end replay. No global pixel comparison or independent confirmation follows.

The tie-complete RDSC variant with canonical track identities TC-RDSC/track-id uses canonical unique track signatures and includes all points within the ninth query-distance radius plus $10^{-12}$; original RDSC remains unchanged. Across 216 inputs/1,080 similarity-row interventions, score failures with absolute and relative score tolerances both $10^{-10}$ are 0 versus original RDSC's 608; neither changes any calibrated answer. Seed changes at the default neighborhood setting (k8) give 0/384 decision switches; varying neighborhood settings k4/k8/k16 across five seeds gives 112 over the full matrix (R14--R15). These are finite sensitivity results, not accuracy gains or universal invariance.

Historical appearance calibration changes BASC-C/No-R/No-Q from ten to seven detections and RDSC from seven to nine when inherited-bank statics replace independent-camera statics (R16). Targets remain the same ten positive conditions; models are partly unavailable and there are no negatives. This constrains calibration transfer without defining a validated replacement recipe.

\subsection{Can learned routing improve the fixed cascade?}
Hold out each of twelve dependence blocks, remove it from calibration of training and target tasks, and preserve own-family exclusion. All 93 full-calibration tasks are predicted once; deletion replicas are not extra training data. Features are camera inheritance, RDSC/transfer/JAPD/edge/raw-q95 margins and RDSC/transfer coverage. Margin is $(s-c)/(|s|+|c|)$ (zero denominator gives zero); transfer first subtracts $1/2$. Missing margin is $-2$. A single-split decision tree (stump) considers training-distinct-value quartile splits, leaves with two blocks, and ordered experts BASC-C/No-Q/RDSC/No-R/recall union. Constants precede splits and win $10^{-12}$ ties. Full split indices, missingness rules and candidate enumeration remain in the exact supporting protocol.

Training task weight is $1/(Gn_g)$ for $G$ groups and $n_g$ tasks in its group. Correct/error/abstention costs are 0/1/0.5; no subsequent sweep is performed. The development-only RDSC-margin split $-0.01860740922501667$ is not an independent result. Table~\ref{tab:learned_results} preserves held-out cohort denominators and comparator losses; equal-block test losses are RDSC 0.331696, BASC-C/best-fixed 0.155159 and stump 0.192212. Pooling all 93 tasks equally gives different losses. The learned selector is not adopted.

\begin{table}[htbp]\centering\small\setlength{\tabcolsep}{3pt}
\begin{tabular}{lrllll}\toprule
Cohort & $n$ & RDSC & BASC-C & Best fixed & Stump\\\midrule
Original legacy & 29 & 0.310 & 0.069 & 0.069 & 0.138\\
Released legacy & 29 & 0.207 & 0.069 & 0.069 & 0.103\\
Original physical & 7 & 0.286 & 0.000 & 0.000 & 0.286\\
Fixed-input physical & 7 & 0.357 & 0.357 & 0.357 & 0.500\\
Regenerated physical & 7 & 0.143 & 0.286 & 0.286 & 0.286\\
Adaptive inherited & 7 & 0.286 & 0.286 & 0.286 & 0.286\\
Fixed inherited & 7 & 0.500 & 0.500 & 0.500 & 0.500\\
\bottomrule\end{tabular}

\caption{Out-of-group mean task losses after group-excluded recalibration, $(FN+FP+0.5A)/n$. Complete counts are separated in Table~\ref{tab:learned_counts}. Cohort denominators sum to 93 dependent tasks; the table does not merge calibration-deletion replicas. Best fixed is selected only on training groups. Physical event labels remain domain-confounded as described in S6.1.}\label{tab:learned_results}
\end{table}
\begin{table}[tbp]\centering\small\setlength{\tabcolsep}{3pt}
\begin{tabular}{lrllll}\toprule
Cohort & $n$ & RDSC & BASC-C & Best fixed & Stump\\\midrule
Original legacy & 29 & 16/3/4/4/2 & 20/6/1/0/2 & 20/6/1/0/2 & 18/6/1/2/2\\
Released legacy & 29 & 19/3/4/1/2 & 20/6/1/0/2 & 20/6/1/0/2 & 19/6/1/1/2\\
Original physical & 7 & 3/2/0/2/0 & 5/2/0/0/0 & 5/2/0/0/0 & 3/2/0/2/0\\
Fixed-input physical & 7 & 3/1/0/2/1 & 3/1/0/2/1 & 3/1/0/2/1 & 2/1/0/3/1\\
Regenerated physical & 7 & 4/2/0/1/0 & 5/0/2/0/0 & 5/0/2/0/0 & 4/1/1/1/0\\
Adaptive inherited & 7 & 4/1/1/1/0 & 4/1/1/1/0 & 4/1/1/1/0 & 4/1/1/1/0\\
Fixed inherited & 7 & 0/0/0/0/7 & 0/0/0/0/7 & 0/0/0/0/7 & 0/0/0/0/7\\
\bottomrule\end{tabular}

\caption{Complete TP/TN/FP/FN/A counts for Table~\ref{tab:learned_results}, preserving all 93 dependent tasks. Equal-block losses in the text use different weights from pooled task losses.}\label{tab:learned_counts}
\end{table}

\subsection{Paired corrections and fallback value}
With separate cameras, the global diagnostic and RDSC are both correct on nine acquisitions. Only global is correct on two, only RDSC on one, and both are wrong on zero. Using this same order, global versus BASC-C gives 10/1/1/0 and RDSC versus BASC-C gives 9/1/2/0. With inherited cameras the three comparisons give 8/3/1/0, 9/2/0/1 and 8/1/1/2, respectively. Every case in which a local method corrected a global error was static. Choosing the correct answer in hindsight gives an upper bound on selector performance.

\paragraph{Abstention cost and fallback value.}
Composition cost is $L(r,\alpha)=(\mathrm{FN}+r\mathrm{FP}+\alpha A)/N$, with all requested labeled non-calibrators in $N$, unknown truth separate and dependent deletion folds retained. Compare fixed methods/constants over $r\in[0.1,10]$, $\alpha\in[0,1]$; constant positive/negative/abstained predictions distinguish refusal cost from accuracy. Complete polygons, ties, coverage, selective errors and per-case records are supplied.

Adaptive-inherited BASC-C/RDSC gives four TP, two TN, one FN; No-R abstains seven times. Fallback strictly improves loss for $\alpha>1/7$, tying at equality. Among all components/constants, BASC-C ties the minimum only for $r\geq0.5,\alpha\geq1/7$, identically to RDSC; no full cohort strictly beats every comparator. Refused physical requests remain refusals even when cheap. These results apply to the specified hypothetical costs.

Direct loss sums, polygon half-planes and a complete cost grid verify these unchanged predictions/priorities. The supported gain is conditional fallback coverage.

\subsection{Complete matched primary local comparisons}
\begin{figure}[tbp]\centering
\includegraphics[width=\linewidth]{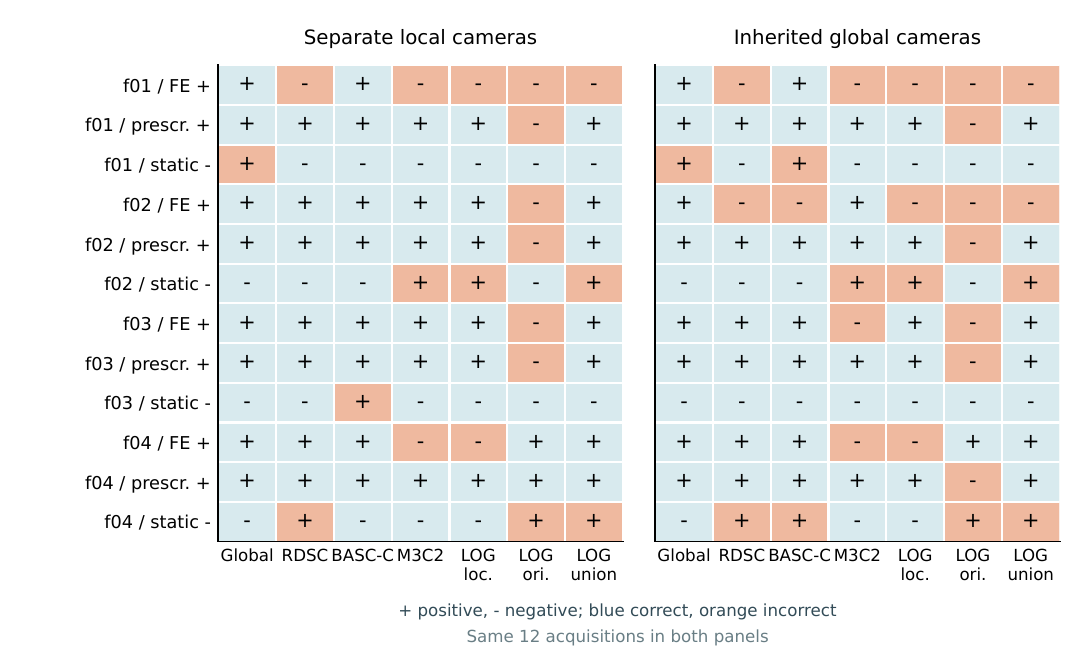}
\caption{Complete core decision matrix. The same twelve acquisitions appear under separate and inherited local-camera estimation. LOG loc./ori./union denote the LOG-lite/ZD location/orientation/union adaptations. A $+$ or $-$ gives a positive or negative screen; blue denotes a correct answer and orange an incorrect answer relative to the supplied acquisition label. The left-hand labels give that class as well. This is descriptive paired accounting; an either-correct oracle is not used for inference.}\label{fig:threecases}\end{figure}

Table~\ref{tab:local_full} and Figure~\ref{fig:threecases} use the EPC input states in S2.1. Transfer, raw median and JAPD match the strongest separate-camera counts. Coherent amplitude matches RDSC in both regimes; ratio detects five positives per regime. LOG-lite/ZD location detects six/five, orientation two/one, and their union seven/six but with a second false alarm. All 216 windows score after the ZD extension; the unextended port retains twelve sequence abstentions per regime. These are adapted-feature outcomes, not measurements of the full M3C2-LOG method.

\begin{table}[tbp]\centering\small\setlength{\tabcolsep}{4pt}
\begin{tabular}{lcc}\toprule Readout & Independent cameras & Inherited cameras\\\midrule
RDSC & 7/3/1/1/0 & 6/3/1/2/0 \\
BASC-C & 8/3/1/0/0 & 7/2/2/1/0 \\
Coherent amplitude $a$ & 7/3/1/1/0 & 6/3/1/2/0 \\
Ratio $a/b$ & 5/3/1/3/0 & 5/3/1/3/0 \\
Point JAPD & 8/3/1/0/0 & 7/3/1/1/0 \\
Raw 3D median (all pairs) & 8/3/1/0/0 & 5/3/1/3/0 \\
M3C2 distance q95 & 6/3/1/2/0 & 5/3/1/3/0 \\
Raw median (M3C2 support) & 8/3/1/0/0 & 6/3/1/2/0 \\
Raw q95 (M3C2 support) & 6/3/1/2/0 & 6/3/1/2/0 \\
LOG-lite/ZD location & 6/3/1/2/0 & 5/3/1/3/0 \\
LOG-lite/ZD orientation & 2/3/1/6/0 & 1/3/1/7/0 \\
LOG-lite/ZD union & 7/2/2/1/0 & 6/2/2/2/0 \\
\bottomrule\end{tabular}

\caption{Complete matched EPC comparison, TP/TN/FP/FN/A, twelve requests per regime. M3C2 and LOG-lite/ZD are named sequence adaptations, not the original methods' physical-significance tests. M3C2-support controls use only its usable points; all-pair controls do not. All methods have equal designated static calibration access.}\label{tab:local_full}
\end{table}

\section{Failed utility criteria and physical input coverage}\label{sec:snegative}
\subsection{RDSC operator and final utility challenge}
The initial and final studies ask whether combining coherent amplitude with its shuffled reference improves on the simpler controls.

\paragraph{Initial design.} The initial generated study uses 984 pairs: 768 field targets, 108 calibrators and 108 nuisance requests. Supports are cube/tube/plane; fields are bend/bulge. Ninety-six stochastic nulls are distinct, while twelve tiny-field targets repeat calibration arrays and cannot be counted as independent negatives. Fourteen methods are prospective and two additive contrasts exploratory. Twenty frozen-generator banks retain the same supports and parameters, not independent anatomy. Full initial and per-bank counts, definitions and seeds remain in R17--R18. Numerical replay does not establish formula-specific utility.

\paragraph{Final design.} The final phase tests complexity justification rather than selecting another score. It reuses the original cube, tube and plane supports, bend/bulge fields, amplitudes 0.002/0.005/0.01/0.02 and error scales 0.001/0.005/0.015/0.03. Each of ten banks has five profiles: independent (I), radial basis function (RBF) error profiles with length parameters 0.12 (S) and 0.45 (L), independent heterogeneity (H), and long correlation plus heterogeneity (CH). For correlated errors, 64 equally spaced ordinal support centers define $K_{ij}=\exp(-\|p_i-c_j\|^2/(2\ell^2))$, normalized to unit row norm. Standard-normal coefficients give a finite low-rank field. Heterogeneous weights range linearly from 0.5 to 2 along support x and are RMS-normalized. XYZ anisotropy is $(0.5,1,2)$ divided by its RMS. These are specified mathematical models, not measured SfM covariance.

With independent shared/global/local innovations $z_s,z_g,z_l$, the errors are $g=0.5\sigma\sqrt{0.5}(z_s+z_g)$ and $l=\sigma\sqrt{0.5}(z_s+z_l)$. The estimates are $x=p+g$ and $y=p+u+v+l$, where $u$ is the injected field and $v$ is the preceding affine nuisance at amplitude 0 or 0.03, alternating replicate parity. Unlike the initial study, $\sigma$ is a marginal coordinate scale before field correlation, without per-realization vector-RMS normalization. There is no exact reference estimate and no tiny-field replay calibration. Each case has a separate innovation seed, 202610239000000 plus its ordinal; shuffle seeds remain 20260904 plus ordinal.

Each bank/support/profile has 32 calibrators (eight per error scale), sixteen separate null targets and 128 field targets: ten banks $\times$ three supports $\times$ five profiles give 26,400 pairs without duplicate input-array file hashes. Each method uses its own maximum-static cutoff on shared calibrators. Crossing all five target/calibration profiles within bank/support closes 108,000 target/definition records and 756,000 answers before evaluation; none abstains. Table~\ref{tab:final_shift} shows matched profiles, R20 all 25 pairs; every finer cell remains machine-readable.

\paragraph{Metric and predefined improvement criterion.} The implemented class-balanced loss is
\begin{equation}
 L_{\rm bal}(\lambda)=\tfrac12\left[
 \frac{FN+A_+}{N_+}+\lambda\frac{FP+A_-}{N_-}\right].
\end{equation}
Here $N_+=3840$ and $N_-=480$ are all requested positive/null targets per matched profile pooled across ten banks; $A_+,A_-$ are class-specific abstentions, penalized as errors. They are zero in this study. With false-negative and false-positive rates denoted by FNR and FPR, respectively, the loss becomes $\tfrac12(\mathrm{FNR}+\lambda\mathrm{FPR})$. Each bank has 384 positives and 48 nulls across the three equally represented supports. Thus pooled class rates equal the means of the ten per-bank class rates; there is no conditioning on successfully scored cases. A bank win requires strictly lower loss at the pooled favorable interval's geometric midpoint; exact ties are not wins. Improvement is comparator loss minus RDSC loss. This factor-$1/2$ scale differs from the composition cost $L(r,\alpha)$ in S6.4, whose denominator is all seven requested targets rather than separate class counts.

The balanced-risk criterion was fixed before scores: a matched profile must favor RDSC over every simple control across a $\lambda$ interval spanning a factor of at least two within $[0.1,10]$, with minimum loss improvement 0.01 at the geometric midpoint and paired wins against each in at least eight of ten banks. \paragraph{Results and interpretation.} All five profiles fail. In CH, the pooled favorable interval $[0.1,0.75)$ has midpoint 0.273861, minimum gain 0.000391, and bank wins 6/5/4/5/6/6 against the ratio, amplitude, subtraction, difference-of-squares, raw median and raw q95 controls, respectively. The practical margin and repetition failures are retained; no threshold or noise profile is revised.

Independent post-closure recount, regenerated ordinal samples and dense-equation checks pass. Exact protocols, all cells and audit receipts remain in the supporting record/scientific capsule; numerical agreement does not pass a scientific utility gate.

\begin{table}[htbp]\centering\small\setlength{\tabcolsep}{3pt}
\begin{tabular}{llrrrrrrr}\toprule
Target & Cal. & RDSC & Amp. & Ratio & $a-b$ & $a^2-b^2$ & Median & q95\\\midrule
I & I & 252/12 & 201/12 & 1898/11 & 1213/11 & 453/10 & 126/14 & 149/11\\
S & S & 127/14 & 135/10 & 356/13 & 131/11 & 132/10 & 110/12 & 135/8\\
L & L & 96/15 & 104/12 & 224/19 & 102/13 & 106/12 & 83/10 & 89/11\\
H & H & 139/10 & 143/13 & 1685/14 & 503/13 & 145/13 & 132/15 & 103/5\\
CH & CH & 138/17 & 124/16 & 137/19 & 135/17 & 130/17 & 126/15 & 135/17\\
\bottomrule\end{tabular}

\caption{Final RDSC utility study under matched target/calibration error profiles. Each cell is TP/FP out of 3,840 field/480 null targets; I/S/L/H/CH are defined in S7.1. The complete 25-profile matrix is R20. No profile meets the predefined utility criteria.}\label{tab:final_shift}
\end{table}

\subsection{Failed nonlinear local-coordinate candidate}
This separate candidate scores nonlinear local-coordinate discrepancies. Its evaluation requests 3,584 manufactured cases across four retained ideal geometries, clean/point-biased/camera-biased/combined profiles, nulls and two positive amplitudes. A common nonlinear point fit supplies discrepancies; nuisance projection and a studentized quadratic alternative are compared with fixed linear-local/raw-pixel controls. The manufactured geometries add no independent anatomy or physical acquisition.

Calibration requires at least 61 usable cases out of 64. Actual usable calibration counts for f01--f04 are 61, 60, 54 and 64, so f02/f03 have unavailable calibration for every method. They remain abstentions rather than zero false-positive rates. Across all cases, 161 common fit failures remain in the requested denominator. Table~\ref{tab:local_failed} gives the complete primary point-reference-bias comparison at amplitude one. The utility gate passes zero of four geometries; no floor, noise draw or target basis is revised to rescue it.

\begin{table}[htbp]\centering\footnotesize\setlength{\tabcolsep}{3pt}
\begin{tabular}{lrrrrl}\toprule
Family & Usable cal./64 & Candidate TP/72 & Raw reference TP/72 & Pooled raw TP/72 & Primary status\\\midrule
f01 & 61 & 4 & 33 & 34 & utility fails\\
f02 & 60 & -- & -- & -- & calibration unavailable\\
f03 & 54 & -- & -- & -- & calibration unavailable\\
f04 & 64 & 0 & 27 & 26 & utility fails\\\bottomrule
\end{tabular}
\caption{Unsuccessful nonlinear local-coordinate candidate. f01 has 63 scored positives and 57 scored nulls (9/7 abstentions among 72/64 requests); f04 scores all 72/64. Candidate and pooled raw comparator have zero FP on those scored primary nulls; the raw-reference f04 control has one FP. All profiles, amplitudes and controls remain in the machine records.}\label{tab:local_failed}\end{table}

Fixed and pooled-refit raw-pixel controls have similar detections, so no large refit benefit is credited. These exposed development outcomes are not independent global-method confirmation; both this candidate and the $F$-specific gate remain failed.

Two singular value decomposition (SVD) interruptions and fourteen completed files are preserved. A frozen continuation reuses completed outputs and same-equation \texttt{gesvd} fallback under unchanged thresholds, closing all 3,584 cases before evaluation. Full attempts/profiles remain in R24's protocol context; numerical repair does not establish utility.

\subsection{Raw phantom coverage failure}
\noindent\textbf{This experiment evaluates input availability, not detector discrimination.}

ColonSplat's real-data archive (Creative Commons Attribution 4.0, CC BY 4.0; Zenodo DOI \href{https://doi.org/10.5281/zenodo.18763383}{10.5281/zenodo.18763383}) specifies nine C3VDv2 actuation recordings~\cite{c3vdv2}. One lacks \texttt{color\_\allowbreak raw} and remains a requested abstention. Eight supply ninety ordinally spaced raw frames each; five older raw C3VD statics add 450 images but are not matched-appearance negatives. Source IDs, hashes, byte counts and mapper options remain in R19's complete protocol context.

The upstream COLMAP 4.2.0 CPU build~\cite{colmap_upstream_420} estimates the \texttt{OPENCV\_FISHEYE} intrinsics in the global model; local cameras would be independently estimated under fixed global intrinsics. This approximates the source omnidirectional camera rather than reproducing verified calibration. No source pose/depth initializes or repairs models. Globals require 72/90 views; fifteen-frame locals require all requested views and matching global registration. All available globals register only 2--4 views, so no locals are attempted. R19 lists every recording and registration outcome.

All 22 methods abstain on all 88 full/delete-one target definitions, with 252 requested window/regime records retained. The primary subset has 22 target/regime requests per method (nine actuation plus two static targets per regime). This is failed input coverage, not TN, zero detector sensitivity or positive physical validation. The native-fisheye profile leaves original pinhole dispatch/statistics unchanged. Pose files are inspected only after prediction closure; units/conventions/synchronization are not independently verified and they are not physical truth. Interrupted orchestration and qualifications remain in the supporting record without changing these outcomes.

\section{Computational resource profiling}\label{sec:sresources}
\subsection{Machine, schedule and recorded quantities}

Profiling begins after SfM reconstruction and measures the resources needed to score the supplied models (Table~\ref{tab:resources_full}). The schedule has sixty primary requests (twelve acquisitions, five strategy/regime combinations) plus twenty f01/f04 static/FE repeats in reverse strategy order. Request $i$ uses central processing unit (CPU) core $8+(i\bmod4)$, balancing four E cores. Host: Intel Core Ultra 9 275HX, Ubuntu 26.04 build, Linux 7.1.5-76070105-generic; core maximum 3300 MHz, one-thread libraries and S9's runtime. Co-scheduling limits latency inference; process CPU is primary.

The 588 allowed inputs are 360 text models, 120 global/separate-local point arrays and 108 inherited arrays. Global/local requests read four/forty files. Median input bytes: 25,023,147.5 (global), 62,303,331 (separate locals), 62,300,836.5 (inherited locals). Only within-request global-response caching is allowed. BASC-C computes all evidence; RDSC assumes existing locals. Loading and scoring exclude startup. Peak resident set size (RSS) includes imports. R0/R1 invalid attempts remain archived/excluded; R2 completes all eighty requests with unchanged outputs.

\texttt{cpu\_\allowbreak seconds} uses \texttt{time.process\_\allowbreak time}; \texttt{loading\_\allowbreak seconds}, \texttt{scoring\_\allowbreak seconds} and their sum \texttt{total\_\allowbreak seconds} use \texttt{time.perf\_\allowbreak counter}. Summary medians use \texttt{cpu\_\allowbreak median}, \texttt{loading\_\allowbreak median}, \texttt{scoring\_\allowbreak median}, \texttt{total\_\allowbreak median}; load/score medians need not sum to median total. Wall medians [ranges], seconds: global 12.994 [12.303,13.533]; RDSC separate 3.722 [3.543,3.887], inherited 3.697 [3.482,3.846]; BASC-C separate 60.481 [57.614,63.532], inherited 54.225 [52.000,56.938]. Output/timing audits pass. Rounded CPU/wall equality does not make the quantities interchangeable; no cross-runtime speed ratio or minimum lazy cost follows.

\begin{table}[tbp]\centering\small\setlength{\tabcolsep}{4pt}
\begin{tabular}{llrrrrr}\toprule
Strategy & Cameras & CPU (s) & Wall (s) & Load (s) & Score (s) & RSS (MiB)\\\midrule
Global modal & none & 12.99 & 12.99 & 0.45 & 12.54 & 337\\
RDSC & separate & 3.72 & 3.72 & 1.17 & 2.56 & 273\\
RDSC & inherited & 3.70 & 3.70 & 1.19 & 2.51 & 272\\
BASC-C reference & separate & 60.48 & 60.48 & 1.16 & 59.31 & 1416\\
BASC-C reference & inherited & 54.22 & 54.22 & 1.15 & 53.07 & 1416\\
\bottomrule\end{tabular}

\caption{Complete profiling quantities. CPU and total wall seconds are separately recorded medians over twelve primary acquisitions per row; load and score are wall seconds. RSS is the maximum process peak including imports. BASC-C computes all evidence and local reconstruction is excluded.}\label{tab:resources_full}
\end{table}

\section{Reproduction and availability}\label{sec:sreproduction}
\begin{table}[tbp]\centering\small
\begin{tabularx}{\linewidth}{p{.37\linewidth}X}\toprule
Reproduction level & Available scope\\\midrule
Compact scores to decisions & Public stored matrices/component records; not model reconstruction\\
Supplied models to scores & Public selected inputs: 32 global requests and 216 local window/regime records\\
Figures / summary tables & Public selected artwork inputs and compact summary tools\\
All RGB to SfM to results & Incomplete; deleted historical RGB and unavailable pipelines\\
All historical comparisons & Limited by private inputs, unlicensed external source and qualified runtimes\\
Cross-backend bitwise identity & Not guaranteed; score tolerances and decision agreement are separate\\\bottomrule
\end{tabularx}
\caption{Public reproduction scope. Model-to-score replay, stored-score replay, figure generation and complete RGB-to-result reconstruction are different tasks.}\label{tab:reproduction_scope}
\end{table}

\subsection{Reproduce primary results from supplied models}
Table~\ref{tab:reproduction_scope} summarizes public access. Software version \ScreeningSourceVersion~\cite{screeningcode} provides the implementations, compact numerical records and reproduction/artwork tools. The source revision is fixed in the bibliography. Clone and install that revision, then run the checks from the repository root:
\begin{lstlisting}
git clone https://github.com/icirauqui/sfm-nonrigidity-screening.git
cd sfm-nonrigidity-screening
git checkout 3674a4856b9b4896969b22f2deeba1cbacbffa0b
python3.12 -m venv .venv
. .venv/bin/activate
python -m pip install .
python tools/check_source.py
python tools/check_layout.py
python -m unittest discover -s tests -v
python tools/check_reference.py --output outputs/compact.json
python tools/summarize_results.py --output outputs/comparison
\end{lstlisting}
The reference check uses stored projected matrices and component scores and writes \texttt{outputs/compact.json}. The summary command writes \texttt{comparison.json} and \texttt{comparison.csv} under \texttt{outputs/comparison}. Evaluation uses labels only after predictions are fixed. To recompute scores from the supplied models, download the inputs and run the replay command below.

The primary/artwork bundles are publicly deposited as dataset version \ScreeningInputVersion~\cite{screeninginputs}, independently of the software version. The bibliography supplies the verified dataset DOI and revision. Its README, license and attribution distinguish the CC BY 4.0 data from software under the GNU General Public License (GPL). Download and verify from a new directory:
\begin{lstlisting}
hf download icirauqui/sfm-nonrigidity-screening-reproduction \
  --type dataset --revision 0621f69360f78f5379afffb872532648d1e6e85a \
  --local-dir outputs/input_download
cd outputs/input_download
shasum -a 256 -c SHA256SUMS
python -m zipfile -t screening-primary-inputs-v0.1.0.zip
python -m zipfile -t screening-artwork-inputs-v0.1.0.zip
python -m zipfile -e screening-primary-inputs-v0.1.0.zip primary
python -m zipfile -e screening-artwork-inputs-v0.1.0.zip artwork
cd ../..
PRIMARY_INPUTS="$(pwd)/outputs/input_download/primary"
ARTWORK_INPUTS="$(pwd)/outputs/input_download/artwork"
\end{lstlisting}
The official Hugging Face command-line interface (\texttt{hf}) downloads fixed revisions; Linux \texttt{sha256sum -c SHA256SUMS} is equivalent. \texttt{provenance/release\_\allowbreak assets.json} supplies sizes, hashes and URLs. After returning to the checkout, the assignments name absolute extracted roots: primary contains \texttt{INPUT\_\allowbreak MANIFEST.json}/\texttt{CONTRACTS.json}; artwork contains \texttt{SHA256.json}.
\begin{lstlisting}
OPENBLAS_NUM_THREADS=1 OMP_NUM_THREADS=1 MKL_NUM_THREADS=1 \
python tools/replay_models.py --inputs "$PRIMARY_INPUTS" \
  --mode all --workers 4 --comparison decisions \
  --output outputs/primary_replay
\end{lstlisting}
Model replay covers 32 global requests and 216 local window/regime records. Workers read explicit inference allowlists; expected scores enter only after closure. Outputs are \texttt{SCORES.json}, \texttt{REPLAY.json}, \texttt{INPUT\_\allowbreak READS.json} and \texttt{NUMERICAL\_\allowbreak DIFFERENCES.json}. Whole-model refusals remain. Display inputs are isolated, with output/input hashes in \texttt{ARTWORK\_\allowbreak RECEIPT.json}. Output directories must be new.

\subsection{Figures and summary tables}
After defining the extracted artwork root above, run:
\begin{lstlisting}
python -m pip install '.[figures]'
MPLCONFIGDIR=outputs/mpl-cache XDG_CACHE_HOME=outputs/cache \
python tools/render_artwork.py --inputs "$ARTWORK_INPUTS" \
  --output outputs/artwork
\end{lstlisting}
The compact summary command above writes comparison tables. The public artwork tools reproduce the original displays. Figure~1 uses the calibration support table and Figure~3 uses the window schedule and decision rules in S3--S4. Dataset version \ScreeningMvaVersion~\cite{screeningmva} supplies the archived campaign code, protocols, results and checks. Its \ScreeningMvaGuide{} verify the archive and reproduce the deposited artwork for main Figure~2 from the seeded perspective fixture and the eleven saved Cambridge gain spectra.

\subsection{Numerical reproducibility limits}
The historical qualified runtime is CPython 3.12.13, NumPy 2.2.6 and SciPy 1.16.1, with image identity \texttt{sha256:88e6523c5\allowbreak 11d52dc76b09d9d7\allowbreak 9946ab377278af56\allowbreak 061d2126727028a8\allowbreak 8ade9bd}. This identity is provenance, not a public container download. Historical same-image global dictionaries are bitwise equal and decisions exact. The source Dockerfile supplies a new build recipe and does not establish that image identity or strict backend equivalence. Global scalar tolerances are relative (rtol) $10^{-7}$ and absolute (atol) $10^{-8}$; local scores retain rtol $10^{-8}$, atol $10^{-10}$. Ranks, statuses and refusal reasons are checked separately. The original matrix audit uses Gram-entry absolute tolerance $10^{-8}+10^{-7}\|G\|_F$.

A different Linux ARM64 backend gives 92 JAPD/edge scores outside strict tolerance (maximum absolute difference $4.674566005391512\times10^{-5}$), with checked global/local decisions unchanged. Original/reorganized kernels agree exactly on the same backend. \texttt{--comparison decisions} retains every discrepancy and may return \texttt{DECISIONS\_\allowbreak MATCH\_\allowbreak NUMERICAL\_\allowbreak REFERENCE\_\allowbreak MISMATCH}: decision reproduction, not strict score parity. \texttt{--comparison strict} preserves the failure criterion; no expected value, tolerance or cutoff is relaxed. See \texttt{docs/numerics.md} for the full backend record. These checks establish neither new validity nor timing equivalence.

\subsection{Historical and private material}
The private scientific capsule \texttt{three\_\allowbreak strategy\_\allowbreak review\_\allowbreak 20261024\_\allowbreak r2.zip}, SHA-256 \texttt{2b43cae16d78c980\allowbreak f3c1e96ea36aa5b1\allowbreak de6c254df4e7d2e6\allowbreak 86bdcb249067cbdf}, retains the original source/runtime receipts, 588 selected model/point files, full historical negative studies and comparator evidence. Its original \texttt{MANIFEST.json}, \texttt{FREEZE.json} records and qualified commands remain unchanged, separately from the public selected distribution. It is not a public download or a blanket-licensed release. Historical replay remains scoped to that capsule and the nested local-method package below.

Native M3C2 additionally requires the separately qualified macOS ARM64 py4dgeo 1.2.0 environment and supplied installation/binary hashes; the source wrapper alone does not establish native-runtime equivalence. The LOG author's unlicensed snapshot is distinct from the public hash-checked external adapter and retained closed results. Source availability does not authorize redistribution of that comparator implementation. The editable-source helper supports offline pdfLaTeX or cached Tectonic; the current editorial build receipt identifies the actual compiler.

\subsubsection{Historical scope and interface}
Historical score-to-decision replay in the private capsule runs from its unchanged nested package:
\begin{lstlisting}
python /path/to/package/legacy/code/reproduce.py /path/to/package/legacy /new/output/path
\end{lstlisting}
Qualified CPython 3.12 standard-library execution reproduces five prediction files, 306 definitions and twelve grouped folds before evaluator recount. CPython 3.9 retains a last-bit learner-loss strict mismatch. The separate matched-model tool can refit 108 inherited point arrays and recompute 216 component requests from eligible models; neither command restores deleted RGB or end-to-end SfM. Original manifests, exact commands, 93 held-out records and schema keys remain in the supporting protocol and capsule README. BASC-C's final field is \texttt{decision}; comparator \texttt{primary} is not the full cascade answer. Numerical replay, same-backend source parity and scientific validation are distinct checks.

\section{Documentary index and display provenance}\label{sec:sindex}
Open Online Resource~2 (\texttt{ESM\_2.zip}) at \texttt{index.html}: all 25 earlier tables R01--R25 retain their cells, captions and exact TeX in \texttt{TABLES.json}. \texttt{protocols.html}/\texttt{protocols\_\allowbreak v4.tex} preserve the preceding protocols; old references resolve within that edition. A current reading map accompanies the unchanged historical retention map.

Main Figures 1 and 3 illustrate the methods and their candidate fields. Main Figure 2 shows the geometric ambiguity and limits of the modal test. Main Figures 4--6 and Figures S1--S5 show acquisition geometry, color images, comparisons and results. Captions give support, transforms and scales. Private capsule reports retain runtime receipts and attempts. S6--S7 report scientific failures. The records add no independent acquisitions and exclude private inputs and unlicensed comparator source.

\section{Exact ambiguity and conditional power}

The fixture illustrates the viewing-ray construction in main Section~4.1 using signed perspective division on 64 points and twelve primary views. Positive ray scales keep the points in front of the camera. The same-state displaced camera and the later returned state test two distinct interventions. An independent check uses rational arithmetic on 132 additional projections. The equality is exact in this latter check. For a point changed by $\gamma\ne1$ on its first ray, the two candidate points and the original centre are collinear. An additional centre outside this line forms distinct rays to the two candidates. Positive depth, simultaneous state and corresponding visibility remain necessary to interpret the image difference.

For the sensitivity calculation, use the linearized static point and camera responses and temporally varying affine columns at the unloaded Cambridge geometry. The smooth library uses quadratic spatial features and slow temporal factors. The localized library uses spatial Gaussian functions, and the rapid library uses faster temporal factors. Columns are orthonormalized under world-displacement RMS before applying the image derivative. Each direction has a gain after nuisance removal and a potentially smaller gain after projection onto the modal test. The fixed test spans the projected smooth-quadratic library.

If $\boldsymbol\epsilon\sim\mathcal N(\mathbf0,\sigma^2\mathbf I)$, then $\mathbf Q^{\mathsf T}\mathbf P\boldsymbol\epsilon/\sigma$ is standard normal in $q$ dimensions because $\mathbf P\mathbf Q=\mathbf Q$ and $\mathbf Q^{\mathsf T}\mathbf Q=\mathbf I_q$. The mean shift along a tested direction of amplitude $a$ has norm $as/\sigma$. This proves the conditional noncentral law in the article. Replacing the linear displacement response with exact perspective differences gives the finite-displacement noncentral law at the same frozen basis. The numerical comparison uses two pixel noise scales (0.3 and 1 pixel), a common bank of 256 noise draws, and amplitude multipliers 0.5, 1 and 2 relative to the linear target. The family gains and the 369 in-range and 225 out-of-range combinations remain separate. The comparisons use fixed geometry and known independent variance, which is different from the empirical calibration on estimated reconstructions.

The external-geometry comparison uses separate projected rest and measured deformed shapes, including weak and full displacement. An optimizer refits the point and camera parameters starting from the measured unloaded geometry with supplied correspondences and intrinsics. Iteration-limit cases are retained among the 66 requests. The synthetic pixel perturbations and conditional projection test provide distinct evidence from the original physical measurements and from any raw-video inference claim.

\section{Attribution and numerical transport}

\begin{table}[!htbp]
\centering
\caption{Original-export factorial and same-track controls. Core columns use different-family static maxima and transfer uses all four native statics. Each score is calibrated separately. The four arms containing point elimination abstain on the same transfer export. All core requests receive scores.}\label{tab:factorial}
\begin{tabular}{lrrr}
\toprule
Responses removed or readout & Core positives / 8 & Static alarms / 4 & Transfer positives / 20\\
\midrule
None & 8 & 1 & 14\\
Affine & 8 & 1 & 3\\
Camera & 8 & 1 & 16\\
Camera + affine & 8 & 1 & 12\\
Point & 8 & 1 & 15\\
Point + affine & 8 & 1 & 19\\
Point + camera & 8 & 1 & 19\\
Point + camera + affine & 8 & 1 & 19\\
Pixel median & 5 & 1 & 4\\
Pixel q95 & 6 & 1 & 9\\
Camera-conditioned Sampson q95 & 8 & 1 & 13\\
\bottomrule
\end{tabular}
\end{table}

The factorial and fitted epipolar tests use the 32 original exports (Table~\ref{tab:factorial}). In the original transfer comparison, the magnitude controls share the modal operator's numerical abstention, and each cutoff uses the three static acquisitions outside the target family. The factorial scores the median and q95 wherever pixel residuals are available and sets the transfer cutoffs using all four static acquisitions. The replay in S2.3 preserves all 340 decisions under the original rules.

The epipolar comparisons share the same 37 scheduled lagged pairs per acquisition. The alternative basis is fixed. Changing which nuisance columns are removed changes the projected alternative as well as the projected residual. The two partial-removal ties are ties in decisions, not a claim about score equivalence or optimality over all bases. The uniform-sampler Progressive-X adaptation fits at least one model on all 1,184 pairs and gives a zero secondary fraction at the sequence quantile. The local sampler fails to fit on all 1,184. The released implementation recovers two groups on its author-supplied example. Two sequential USAC--MAGSAC fits also give zero secondary fraction at the sequence readout.

For rank-aware elimination, write the fixed-camera track objective as $\|\mathbf e_i-\mathbf B_i\delta\mathbf c-\mathbf A_i\delta\mathbf x_i\|^2$. Minimizing over $\delta\mathbf x_i$ removes exactly the component in the retained column space of $\mathbf A_i$. The construction applies the same orthogonal complement to residual, camera and field columns. The effective degrees of freedom subtract the retained block ranks, rather than three coordinates per track by convention. Downstream rank and variance checks still apply. The streamed/dense mixed-rank fixture differs by at most $7.11\times10^{-15}$, and full-rank replay agrees exactly on the 31 originally scored exports. The two newly scored StereoMIS clips have modal energies of 47,593.78 and 731.04 pixel$^2$ against 89.50 pixel$^2$ in the original-export static calibration. One transfer export still has insufficient residual variance. The original refusal reason remains recorded for the baseline.

With fixed poses and points, the focal term reduces the largest modal energy change under $-2\%$ and $+2\%$ focal perturbations from 165.59 and 110.61 to $3.66\times10^{-5}$ and $6.48\times10^{-5}$ pixel$^2$, respectively. With pixel coordinates and model projection both scaled by $k>0$, raw energy scales by $k^2$ and studentized energy is invariant. Dividing by squared focal length and observation count changes units and exposure to support, but does not correct changes in visibility or a misspecified focal estimate. The every-other-track check and its observed decisions are supplied in Online Resource~3.

In the refit check, the retained image-to-point index is reconstructed without creating new correspondences. Twelve models at each of three focal settings (0.98, 1.00 and 1.02) undergo bounded point-and-camera reoptimization. Both native-scale arms report 8/8 positives and 1/4 static false alarms. The low/high raw arms give 4/4 and 3/4 static false alarms, and the corresponding focal-nuisance arms give 1/4 each. The largest corrected deviations from the native gain are 106,617.03 and 9.86 pixel$^2$, both on f04 non-FE. The larger comes from a fit ending at the iteration limit. All perturbed arms still identify eight positives on this exposed cohort.

\section{Real video, comparison methods and physical extent}

\begin{table}[!htbp]\centering
\caption{Real-video tests with the original exposed synthetic decision threshold. Positives refer to the indicated event or scene-motion labels. Depth is used for a separate evaluation, where available.}\label{tab:real_video}
\begin{tabularx}{\linewidth}{Xrrrl}
\toprule
Requests & Scored & Positives found & Static false alarms & Unscored requests\\
\midrule
Bonn 4 moving / 2 static & 5/6 & 4/4 & 1/2 & 1 static\\
TUM 2 moving / 2 static & 4/4 & 2/2 & 2/2 & 0\\
StereoMIS 7 event requests, rank-aware & 5/7 & 5/7 & -- & 2 event requests\\
D4D 6 clips & 0/6 & -- & -- & 6 clips\\
\bottomrule
\end{tabularx}
\end{table}

Table~\ref{tab:real_video} reports sequence decisions. For the four TUM Freiburg~3 recordings, the registered tracks were scaled to the first ten depth-qualified views and evaluated on later views. The two statics have mean absolute errors of 3.76 and 3.28 cm, with 89.9\% and 93.2\% of errors within 5 cm. The two walking scenes have 20.70 and 25.50 cm errors, with 58.5\% and 54.7\% within 5 cm. The one scored Bonn static has 20.03 cm error, between the best and worst moving recordings. The uncorrected angular score ranks both TUM walking clips above both statics. The focal-corrected angular score does so for one walking clip and flags all four recordings with either the synthetic static bound or the eligible Bonn static threshold. Results are reported separately for these small cohorts.

RoMo and GeoMotion share the 16-frame Bonn inputs. For balloon, crowd, moving box and person tracking, median moving fractions are 0.127, 0, 0 and 0.054 for RoMo, and 0.154, 0.381, 0.223 and 0.132 for GeoMotion. RoMo produces fallback masks for both statics without any trusted frames. GeoMotion returns 0.284 and 0.160 for the statics. These video scores measure moving regions. In the separate orthographic pickup example, the Octave adaptation of TSM-NRSfM~\cite{tsm} has RMS error 0.611 on 135 held-out coordinate pairs in archive units. The authors' completion stage alone gives 0.213, a regularized rank-three fit 0.442, and an unregularized rank-three fit 109.642 (median 0.413). The native MATLAB route requires an unavailable semidefinite programming (SDP) dependency. The observation models differ from the perspective score.

The DRENDS depth evaluation uses 44 colon and 46 intestine frames with synchronized range measurements. Dense stereo has mean absolute errors of 48.83 and 57.99 mm at reference coverage of 6.5\% and 7.6\%. Privileged persistence gives 5.87 and 13.16 mm on available intersections and remains better on common support. Sparse SuperPoint/LightGlue gives 24.81 and 33.11 mm on 1,028 and 679 reference-matched points. The correction learned on eight early colon frames gives 33.34 and 37.63 mm at 6.4\% and 7.6\% coverage, missing the 10-mm and 50\% feasibility criteria.

D4D supplies eight unique structured-light endpoint clouds and 326 eligible monocular frames from two sessions of one specimen. Three exhaustive rematching requests time out at 1,200 s. Two register 3/42 and 3/51 views, below the 80\% requirement, and the sixth has too few source frames. Over twelve optical endpoints, semi-global block matching (SGBM) covers 84.8--87.3\% of the reference raster with errors of 3.02--9.43 mm. The released stereo depths cover the clipped raster with errors of 2.83--11.04 mm. Their endpoint change errors are 0.49--10.93 mm, and zero change gives a smaller error in five of six clips. Stereo change errors are 0.57--0.72 mm for four repeated-state pairs. The optical reference uses an independent sensor, with endpoint delays and privileged registration.

\section{Synchronized view choices and reference support}

\begin{table}[!htbp]\centering
\caption{Synthetic view-choice replay. Each row contains 720 displaced requests. The same 240 geometry/noise seeds appear in each regime. Success means reconstruction RMS no more than 0.025 scene units. Fixed is the retrospectively best single action in the row.}\label{tab:synthetic_view_choice}
\begin{tabular}{lrrrr}
\toprule
Visibility, timing, bias & Fixed & Uncertainty & Targeted & Oracle\\
\midrule
Open, synchronous, unbiased & 720 & 720 & 720 & 720\\
Open, synchronous, biased & 720 & 472 & 472 & 720\\
Open, late, unbiased & 479 & 479 & 479 & 480\\
Open, late, biased & 470 & 262 & 262 & 478\\
Occluded, synchronous, unbiased & 720 & 720 & 712 & 720\\
Occluded, synchronous, biased & 720 & 117 & 704 & 720\\
Occluded, late, unbiased & 466 & 398 & 466 & 467\\
Occluded, late, biased & 466 & 72 & 457 & 466\\
\bottomrule
\end{tabular}
\end{table}

The synthetic selectors share the same static reference, calibrated poses and supplied correspondences (Table~\ref{tab:synthetic_view_choice}). The fixed action along the $y$ axis is geometrically advantaged under the imposed $x$-axis bias. The oracle has at most 1.1 percentage points of headroom against the retrospectively best fixed action, below the 10-point criterion. Accurately reconstructed static and moving geometry under a tolerance is the outcome, distinct from correct motion labels.

\begin{table}[!htbp]\centering
\caption{SceneHub4D synchronized SuperPoint/LightGlue replay. Counts meet 10 cm against independently measured anchor depth. One additional view is available per decision. Oracle uses both measured action outcomes retrospectively. Anchor features without a reliable reconstruction remain in the denominator.}\label{tab:scenehub_view_choice}
\begin{tabular}{lrrrrr}
\toprule
Recording & Sensor requests & Fixed 1 & Fixed 2 & Adaptive & Oracle\\
\midrule
Couch, static & 5,483 & 3 & 0 & 0 & 3\\
Couch, dynamic & 8,083 & 0 & 0 & 0 & 0\\
Kitchen, static & 9,654 & 2,267 & 49 & 49 & 2,267\\
Kitchen, dynamic & 10,367 & 2,233 & 23 & 1,587 & 2,233\\
Kitchen, changed region & 720 & 137 & 0 & 106 & 137\\
\bottomrule
\end{tabular}
\end{table}

The SceneHub4D replay offers two candidate cameras at each of the 38 instants and reserves the anchor camera's infrared depth for evaluation (Table~\ref{tab:scenehub_view_choice}). The attempted SIFT prior has too little support and defaults to the maximum-separation rule. The learned-feature prior has 400 kitchen and three couch reference points. The changed-region row is a subset of the dynamic kitchen recording. Repeated instants and feature requests are not independent scenes, and sparse successful points do not establish complete-surface utility.

In the seven-view Cambridge physical rig, the replay audits 25,830 view partitions over 123 pressure states in eleven configurations. At most four common markers have two observed rays on both sides of any disjoint 3+4 split per state. Pooling all cameras gives at least two rays for all 15 markers in 109 of 123 states. The disjoint-view restriction bounds independent coverage for this design. It is distinct from the controlled perspective reprojections in S11.

Online Resource~3 contains the 24 November protocols, records, numerical summaries and analysis code. Provider data, large outputs and learned weights are retrieved separately under their respective terms.

\small
\raggedright